%% file: main.tex
\documentclass[twoside,11pt]{article}

\usepackage{blindtext}

\usepackage[abbrvbib, preprint]{jmlr2e}
\usepackage{tcolorbox}
\usepackage{amsmath}
\usepackage{booktabs}

\usepackage{lastpage}

\ShortHeadings{Admissibility of Stein Shrinkage for BN in the Presence of Adversarial Attacks}{Ivolgina, Fletcher and Vemuri}
\firstpageno{1}

\begin{document}

\title{Admissibility of Stein Shrinkage for Batch Normalization in the Presence of Adversarial Attacks}

\author{\name Sofia Ivolgina \email sivolgina@ufl.edu \\
       \addr Department of Statistics\\
       University of Florida\\
       Gainesville, Fl 32611 USA
       \AND
       \name P. Thomas Fletcher \email ptf8v@virginia.edu \\
       \addr Department of Electrical and Computer Engineering\\
       University of Virginia\\
       Charlottesville, VA 22904, USA
       \AND 
       \name Baba C. Vemuri \email vemuri@ufl.edu \\
       \addr Department of Comput. Info. Sci. \& Eng. \\
       University of Florida\\ Gainesville, Fl 32608, USA}

\editor{My Editor}

\maketitle

\input{abstract}

\begin{keywords}
James-Stein shrinkage estimator, Admissibility,  Batch normalization, Sub-Gaussian random variables, Adversarial attacks
\end{keywords}

\input{intro}

\input{SteinShrinkage}
\input{proof}

\input{experiments}

\input{conc}


\acks{
This research was supported in part by NIH National Institute of Neurological Disorders (NINDS) and National Institute for Aging (NIA) under Grant RO1NS121099 to Vemuri. It was also partially supported by the NSF Smart and Connected Health grant 2205417 to Fletcher.

{\bf PPMI Data Availability Statement:}
Data used in the preparation of this article were obtained from the Parkinson’s Progression Markers Initiative (PPMI) database [(https://www.ppmi-info.org/access-dataspecimens/download-data, accessed on 6 December 2023, RRID:SCR 006431]. For up-to-date information on the study, visit www.ppmi-info.org. This analysis used data openly available from PPMI, obtained from PPMI upon request.

{\bf PPMI Data Acknowledgments:} We extend our gratitude to Lesa Melnyczuk and Iryna Milne for their valuable assistance. Data used in the preparation of this article were obtained on 6 December, 2023 from the Parkinson’s Progression Markers Initiative (PPMI) database (https://www.ppmi-info.org/access-data-specimens/download-data, accessed on 6 Decemeber 2023), RRID:SCR-006431. For up-to-date information on the study, visit http://www.ppmi-info.org, accessed on 6 Dec. 2023. PPMI—a public–private partnership—is funded by the Michael J. Fox Foundation for Parkinson’s Research and funding partners, including 4D Pharma, Abbvie, AcureX, Allergan, Amathus Therapeutics, Aligning Science Across Parkinson’s, AskBio, Avid Radiopharmaceuticals, BIAL, BioArctic, Biogen, Biohaven, BioLegend, BlueRock Therapeutics, Bristol-Myers Squibb, Calico Labs, Capsida Biotherapeutics, Celgene, Cerevel Therapeutics, Coave Therapeutics, DaCapo Brainscience, Denali, Edmond J. Safra Foundation, Eli Lilly, Gain Therapeutics, GE HealthCare, Genentech, GSK, Golub Capital, Handl Therapeutics, Insitro, Jazz Pharmaceuticals, Johnson \& Johnson Innovative Medicine, Lundbeck, Merck, Meso Scale Discovery, Mission Therapeutics, Neurocrine Biosciences, Neuron23, Neuropore, Pfizer, Piramal, Prevail Therapeutics, Roche, Sanofi, Servier, Sun Pharma Advanced Research Company, Takeda, Teva, UCB, Vanqua Bio, Verily, Voyager Therapeutics, the Weston Family Foundation, and Yumanity Therapeutics.
}

\newpage

\newpage
\appendix
\input{appendix}
\vskip 0.2in
\bibliography{references}

\end{document}

%% file: abstract.tex
\begin{abstract}
Batch normalization (BN) is a ubiquitous operation in deep neural networks, primarily used to improve stability and regularization during training. BN centers and scales feature maps using sample means and variances, which are naturally suited for Stein's shrinkage estimation. Applying such shrinkage yields more accurate mean and variance estimates of the batch in the mean-squared-error sense.  In this paper, we prove that the Stein shrinkage estimator of the mean and variance dominates over the sample mean and variance estimators, respectively, in the presence of adversarial attacks modeled using sub-Gaussian distributions. Furthermore, by construction, the James–Stein (JS) BN yields a smaller local Lipschitz constant compared to the vanilla BN, implying better regularity properties and potentially improved robustness. This facilitates and justifies the application of Stein shrinkage to estimate the mean and variance parameters in BN and the use of it in image classification and segmentation tasks with and without adversarial attacks.
We present SOTA performance results using this Stein-corrected BN in a standard ResNet architecture applied to the task of image classification using CIFAR-10 data, 3D CNN on PPMI (neuroimaging) data, and image segmentation using HRNet on Cityscape data with and without adversarial attacks.
\end{abstract}

%% file: intro.tex
\section{Introduction}
\label{sec:intro}
Deep Neural Networks (DNNs) are now widely accepted and used as the primary workhorse in many of the complicated function approximation tasks encountered in engineering and science. In particular, the field of computer vision has seen a major paradigm shift from traditional innovative feature engineering to data-driven feature learning via the use of DNNs. DNN training is well known to be time intensive and unstable. Batch normalization (BN) is a particularly useful tool in achieving regularization and numerical stability  during network training. BN typically involves centering and scaling feature maps. The centering and scaling operations involve computing the sample means (SMs) and variances (SVs) from the samples/members in the batch, respectively. This task is ideally suited for the application of Stein's shrinkage method, which is well known in the field of statistics. Before delving into Stein's shrinkage method, let us briefly review the widely popular BN in DNNs. 

BN was introduced in the machine learning literature primarily to accelerate training and reduce the so-called internal covariate shift \citep{Ioffe-ICML15}. Since its inception, several variants have populated the literature, including, but not limited to, layer normalization in Recurrent Neural Networks (RNNs) \citep{ba2016layernormalization} and Transformers, spectral normalization in generative adversarial networks (GANs) \citep{Miyato-ICLR18}, and many others. Improvements in BN in the form of a stochastic ``whitening'' transform to decorrelate the features in each layer at each step of the training have also been proposed in \cite{Zhang_2021_CVPR} and BN preconditioning in \cite{Lange-JMLR2022}, wherein the preconditioner is constructed using batch statistics.  These normalization algorithms do not address the effectiveness of theoretical improvements in the stability or convergence of training in the presence of adversarial attacks.

The primary motivation of this paper is to explore whether classical statistical techniques, specifically James–Stein (JS) shrinkage estimators, can bring theoretical and practical benefits when applied to DNNs. Although many advances in adversarial robustness focus on architectural changes or alternative training objectives, our approach is fundamentally different: we revisit a classical estimator from multivariate statistics and investigate how its well-understood risk-reduction properties behave in the presence of adversarial perturbations.

This motivation comes from previous empirical work by \cite{khoshsirat2023improvingnormalizationjamessteinestimator}, which demonstrated that applying shrinkage estimators to the mean and variance of BN layers improved performance under standard non-adversarial conditions. The improvement can be attributed to the fact that shrinkage estimators reduce the mean squared error (MSE) under squared loss when estimating multiple parameters jointly -- a setting that naturally arises in BN, where each layer must estimate a vector of means and variances simultaneously.

In this work, we extend this line of inquiry to adversarial scenarios. Specifically, we address the problem of designing a variant of BN that maintains or improves the prediction accuracy in the presence of adversarial attacks, modeled as additive bounded perturbations \citep{MauryaArxiv22}. Popular attack methods satisfying this bounded perturbation assumption include the fast gradient sign method (FGSM) \citep{goodfellow2015explaining} and the iterative variant projected gradient descent (PGD). Unlike many existing defenses that focus on robust losses or architectural modifications, 
{\it our goal here is not the development of a statistically robust estimator that is insensitive to adversarial attacks by seeking a robust loss or modifying the network architecture, but rather, to adapt an existing shrinkage estimator, specifically, the well known JS-shrinkage estimator and prove that it retains its dominance over the SM and SV in the presence of adversarial perturbations. We theoretically show and empirically confirm that this dominance translates into more stable BN statistics, which in turn are observed to yield improved robustness and prediction accuracy.}


\subsection{A Brief Note on the JS-shrinkage Estimator}
In the following we present a brief introduction to the JS-shrinkage estimator and motivate its aptness in the context of BN in DNNs. This will provide the necessary background and motivation for our work presented in this paper.

Shrinkage estimation is a parameter estimation technique that is primarily used for variance reduction and to improve parameter estimation accuracy by shrinking the parameter towards a fixed target. For instance, shrinking the estimate of the mean of a Gaussian toward zero. It is especially useful in high-dimensional and scarce data settings. The shrinkage, of course, leads to a biased estimate, but with smaller variance, which is sought after in many applications. 
The JS-shrinkage estimator was proposed in the field of statistics as an estimator of the mean for samples drawn from a Gaussian distribution and has been shown to dominate the maximum likelihood estimator (MLE) in terms of the risk (mean-squared error) \citep{James-Stein61}. 
Here, \textit{dominance} is understood in the general decision-theoretic sense. 
Suppose we have a random variable $x$ following some distribution $p(x|\theta)$ 
and we wish to estimate a parameter $\theta$. 
Let the \textbf{risk function} be $R(\hat{\theta}, \theta) = \mathbb{E}\!\left[ L(\hat{\theta}, \theta) \right],$ where $L$ is a loss function and $\mathbb{E}$ is the expectation operator. In our setting, we use the \textit{squared error loss}
$L(\hat{\theta}, \theta) = (\hat{\theta} - \theta)^2.$
We say that an estimator $\hat{\theta}_1$ \textit{dominates} another estimator $\hat{\theta}_2$ if $R(\hat{\theta}_1, \theta) \leq R(\hat{\theta}_2, \theta) \quad \text{for all } \theta \in \Theta,$ and there exists at least one $\theta_0 \in \Theta$ such that
$R(\hat{\theta}_1, \theta_0) < R(\hat{\theta}_2, \theta_0)$
\citep[Chapter~5]{lehmann1998point}.

In the case of the JS-estimator, the parameter $\theta$ 
is a $p$-dimensional mean vector $\boldsymbol{\mu}$ of a Gaussian random vector $\mathbf{X} \sim N(\boldsymbol{\mu}, \mathbf{I})$. 
Under squared error loss, the JS estimator $\boldsymbol{\mu}^{JS}$ dominates the maximum likelihood estimator (MLE) $\boldsymbol{\mu}^{MLE}$, which means that
\[
\mathbb{E}\!\left[ \| \boldsymbol{\mu}^{JS} - \boldsymbol{\mu} \|^2 \right] 
\ \leq\ 
\mathbb{E}\!\left[ \| \boldsymbol{\mu}^{MLE} - \boldsymbol{\mu} \|^2 \right],
\]
with strict inequality for some $\boldsymbol{\mu}$. 
Importantly, this improvement concerns the \textit{overall} estimation risk of the mean vector, not the risk of individual components.

At the foundation of the JS-shrinkage estimator lies Stein's lemma (also known as Stein's identity), which characterizes the normal distribution. This lemma is also fundamental to Stein's method, widely used for distributional comparisons. We will not discuss this well-known method here and instead refer the interested reader to the recent comprehensive survey by \cite{Anastasiou23}.

Since its inception, the JS shrinkage estimator has been generalized and new optimal shrinkage estimators have been developed for application to many types of distributions such as the location family \citep{Brown66}, exponential family \citep{Hudson78AOS}, gamma distribution \citep{DasGupta86AOS}, distributions with quadratic variance function \citep{XianchaoXie-Brown-AOS16} and work by \cite{FathiAOS2022} to relax the Gaussian assumption in high dimensions. For a comprehensive survey on Stein's method and shrinkage estimators, we refer the reader to \citep{Kubokawa24}. In addition, more recently, Stein's shrinkage estimate has been generalized to the multiple mean estimation problem in the context of kernel mean embeddings using convex combinations of empirical mean estimators \citep{blanchard2024estimation}, and to distributions on manifold-valued random variables \citep{Yang-Doss-Vemuri22}.

\subsection{Relevance of JS-estimator to BN}





Given the above brief introduction to the JS-shrinkage estimator, we are now ready to present it's relevance to BN in DNNs and how it can be applied to improve the stability and accuracy of predictions. 
A wide range of normalization techniques have been developed to facilitate the training of DNNs by normalizing intermediate feature distributions and improving the stochastic optimization dynamics. One of the earliest and most influential approaches is BN, introduced in \cite{Ioffe-ICML15}, which normalizes intermediate features using statistics computed over a mini-batch, estimating for each channel both a mean and a variance. In the following, we give a brief and precise description of different types of normalization that are in vogue in the literature.

Let $X \in \mathbb{R}^{N \times C \times H \times W}$ be the feature tensor, where $N$ is the batch size, $C$ the number of channels (feature maps), and $H, W$ the spatial dimensions. 
Many normalization schemes follow the same basic operation:
\[
\hat{x}_i = \frac{x_i - \mu_i}{\sqrt{\sigma^2_i + \epsilon}}, \quad 
\mu_i = \frac{1}{|S_i|}\sum_{k \in S_i} x_k, \quad
\sigma_i^2 = \frac{1}{|S_i|}\sum_{k \in S_i} (x_k - \mu_i)^2,
\]
where $S_i$ is the subset of tensor elements over which the statistics are computed. 
The choice of $S_i$ defines the normalization type: 
BN averages over the batch and spatial positions (per channel) \citep{Ioffe-ICML15}, 
LayerNorm averages over channels and spatial positions (per sample) \citep{ba2016layernormalization}, 
InstanceNorm averages over spatial positions (per sample, per channel) \citep{ulyanov2017instancenormalizationmissingingredient}, 
and GroupNorm averages over spatial positions and a group of channels \citep{wu2018groupnormalization}.
After normalization, a learnable affine transformation is applied to preserve the representational flexibility:
$y_i = \gamma_i \hat{x_i} + \beta_i,$ where $\gamma_i, \beta_i$ are trainable parameters indexed by coordinates across which we compute the statistics.

Despite the development of alternatives less sensitive to batch size, BN remains widely used due to its strong empirical impact on optimization dynamics \citep{conf/nips/SanturkarTIM18} and generalization. 
A key observation is that BN estimates a large number of mean and variance values simultaneously, one pair for each channel. This can be naturally viewed as a multitask estimation problem: we estimate a vector of means $\mu = (\mu_1, \dots, \mu_c)$ and variances $\sigma^2 = (\sigma^2_1, \dots \sigma^2_c)$, each based on a finite number of observations. 

In such settings, JS-shrinkage estimators offer a statistically grounded way to improve estimation by sharing information across tasks, in this case, across channels (feature maps). The classical result suggests that when estimating multiple parameters jointly under squared loss, shrinking individual estimates toward their grand mean can reduce overall error. Applying this idea in the context of BN could lead to more accurate and stable normalization statistics, particularly in small batch or high variance regimes, where the empirical estimates of mean and variance are noisy and benefit from regularization.

Having established adequate motivation for application of Stein's shrinkage to BN in a DNN, we are now ready to narrate the rest of the story. But before that, a word on what is to follow. Section \ref{SShrink} delves into the presentation of the appropriate shrinkage formulas for the BN. This is followed by the presentation of theoretical results on the dominance of Stein's shrinkage formulas in the presence of adversarial attacks in Section \ref{Dominance}. In Section \ref{expts}, we present experimental results that compare the performance of the proposed shrinkage to SOTA on several benchmark data sets. Finally, in Section \ref{DC}, we present the discussion and conclusions.

%% file: SteinShrinkage.tex
\section{Stein's Shrinkage Applied to BN}\label{SShrink}

In this section, we apply Stein's shrinkage to the BN and present formulas for it's computation in DNNs. 

Let $X \sim \mathcal{N}_p(\theta, I_p)$, where $\theta \in \mathbb{R}^p$ is an unknown parameter vector and $I_p$ is the $p$-dimensional identity matrix. 
JS-shrinkage estimator of the unknown parameter $\theta$ is given by
\[
\hat{\theta}^{\mathrm{JS}} = \left(1 - \frac{p - 2}{\|X\|^2_2} \right) X.
\]
It is well known that JS-Estimator has a smaller mean-squared error than the maximum likelihood estimator (MLE) but is obviously a biased estimator. 
However, a smaller mean squared error is desirable in many applications. Most recently, JS-estimator for the computation of the mean in BN was used in \cite{khoshsirat2023improvingnormalizationjamessteinestimator}. However, the authors in \cite{khoshsirat2023improvingnormalizationjamessteinestimator} used the same JS-estimator formula to shrink the mean as well as the variance in BN. Directly applying the same shrinkage formula to the variance parameter is, however, theoretically flawed, as the SV is not normally distributed but instead follows a scaled chi-squared distribution, a special case of the Gamma distribution. To address this, we adopt a principled alternative.

Much of the theoretical work on BN implicitly relies on the assumption that activation maps that form the input to normalization layers are approximately Gaussian \citep{lee2018deep, Neal:1994a, Neal:1994b}. This Gaussianity assumption enables analytical tractability and underpins many prior analyses of BN dynamics. 
Following this, we assume that the individual elements of the feature tensor in BN are approximately normally distributed. 
Under this assumption, the SM is normally distributed, while the SV follows a chi-squared distribution. Specifically, for a given channel with $n = N \cdot H \cdot W$ observations, 
$\hat{\sigma}_i^2 \sim (\frac{\sigma_i^2}{n} \chi^2_{n-1}) \sim \mathrm{Gamma}\left( \alpha = \frac{n - 1}{2},\ \beta_i = \frac{2\sigma_i^2}{n} \right)$.
Here, $\chi^2$ denotes the $\chi$-squared distribution, which is a special case of the well known Gamma distribution with shape parameter $\alpha$ and scale parameter $\beta$. 
This enables us to apply the shrinkage to the scale parameter of the Gamma distribution using the Stein shrinkage estimator presented in \cite{DasGupta86AOS}.

Let $X_i \sim \mathrm{Gamma}(\alpha, \beta_i)$ be independently distributed for $i = 1, \dots, p$, where all shape parameters are equal: $\alpha_1 = \dots = \alpha_p = \alpha$. 
Then, JS-estimators for the scale parameters $\beta = (\beta_1, \dots, \beta_p) \in \mathbb{R}_+^p$ are as follows:
\begin{equation}
\hat{\beta}_i^{\,\mathrm{JS}} =\frac{X_i}{\alpha + 1} + \tilde{c}V, \quad 
\text{where} \quad V = \left( \prod_{j=1}^p X_j \right)^{1/p}, \quad 
\tilde{c} \in \left[0, \dfrac{2(p - 1)}{(\alpha + 1)(\alpha p + 1)} \right].
\label{gamma_shrinkage}
\end{equation}

To incorporate the JS correction into BN, we compute the statistics for each channel and apply the shrinkage formulas as outlined below. 
Given a 4D tensor input $x \in \mathbb{R}^{N\times C \times H \times W},$ we define $\mu_C \in \mathbb{R}^C$ and $\sigma^2_C \in \mathbb{R}^C_{+}$ as the channel-wise mean and variance computed over all spatial and batch dimensions:
\[
\mu_C = \frac{1}{N H W}\sum_{n, h, w} x_{n, c,h,w}, \quad 
\sigma^2_C = \frac{1}{N H W}\sum_{n, h, w} (x_{n, c,h,w} - \mu_C)^2.
\]
We then apply shrinkage to these estimates. The JS-corrected mean and variance are:
\begin{tcolorbox}[colback=gray!5!white,colframe=gray!75!black,title=Stein Shrinkage Forumals for Mean \& Variance]
\begin{equation}
\begin{aligned}
  \mu^{\mathrm{JS}}_C  &= \left(1 - \frac{(c - 2) \sigma^2_{\mu_C}}{\|\mu_C\|^2_2} \right) \mu_C, \quad {\sigma_C^2}^{\mathrm{JS}}= \frac{n}{n + 1} \hat{\sigma}_i^2 + \tilde{c} \cdot V, \\
  &\text{where } V = \left( \prod_{j=1}^p \hat{\sigma}_j^2 \right)^{1/p},\quad 
  \tilde{c} \in \left[0, \dfrac{4n(p - 1)}{(n + 1)((n-1) p + 2)} \right].
\end{aligned}
\label{JS-Corrections}
\end{equation}
\end{tcolorbox}
Here, $\sigma^2_{\mu_C}$ denotes the estimated variance of the corresponding vector $\mu_C$.

Finally, the normalized output is given by: 

\begin{tcolorbox}[colback=gray!5!white,colframe=gray!75!black,title=Final Normalized (scaled and shifted) Output]
\begin{equation} 
\hat{x} = \frac{x - \mu^{\mathrm{JS}}}{\sqrt{{\sigma^2}^{\mathrm{JS}} + \epsilon}}, \quad y = \gamma \hat{x} + \beta.
\label{finalCorrection}
\end{equation}
\end{tcolorbox}
Equations \ref{JS-Corrections} and \ref{finalCorrection}, respectively, are the formulas we advocate for use in BN corrections and will be used in all our experiments.

We would like to mention that the choice of the shrinkage constant can, in principle, influence the estimator. When the constant approaches zero, the estimator reduces to the empirical variance, so values near the boundary of the admissible interval are of primary theoretical interest. 
In our experiments, we did not observe a significant sensitivity to moderate changes in this constant. 
Following the precedent set in the original work on shrinkage estimators for the gamma distribution \citep{DasGupta86AOS}, we adopted boundary values of the shrinkage constant in our experiments, as detailed in Section~\ref{expts}. We are now ready to present the theoretical results of the dominance of these shrinkage estimators in the presence of adversarial attacks.

%% file: proof.tex
\section{Dominance of Stein Shrinkage in the Presence of Adversarial Attacks} \label{Dominance}

In this section, we present the main theoretical contribution of our work namely, that the JS-estimator continues to exhibit its dominance even under adversarial perturbations. 
We begin by formalizing how adversarial attacks can be modeled probabilistically. Then, we introduce the necessary preliminaries and outline the main ideas behind the proof and include the detailed proof in the Appendix \ref{Appendix}.

We model adversarial perturbations using a sub-Gaussian distribution, which is well motivated both practically and theoretically. Adversarial perturbations are typically bounded in magnitude (for example, under an $l_p$-norm constraint) to ensure that the model remains learnable - a standard assumption in the literature \citep{cohen2019certifiedadversarialrobustnessrandomized, zhai2019adversariallyrobustgeneralizationjust}. Since bounded random variables are inherently sub-Gaussian, this provides a natural modeling choice. Additionally, the sub-Gaussian model enables a tractable analysis via the use of concentration inequalities and allows one to work with a well-defined variance proxy. 

Following \citep{MauryaArxiv22}, we assume that adversarial perturbations $Y \in \mathbb{R}^p$ are mean-zero sub-Gaussian with variance proxy $2\varepsilon^2$, i.e., $\mathbb{E}\!\left[ e^{\lambda \langle v, Y \rangle} \right] \leq e^{\varepsilon^2 \lambda^2},\ \forall \lambda \in \mathbb{R}, \ \forall v \in \mathbb{R}^p: \|v\|=1,$ which we denote compactly as $Y \sim \mathrm{SG}(2\varepsilon^2)$.
This assumption allows us to model the observed vector as a sum of noiseless data and adversarial noise:
$Z = X + Y, \quad X \sim \mathcal{N}_p(\theta, \sigma^2 I).$


\subsection{Dominance of the JS-estimator under Sub-Gaussian Perturbations for Normally Distributed Random Variables}
Consider the problem of estimating a parameter vector $\theta \in \mathbb{R}^p$ based on corrupted observations $Z=X+ Y,$ where $X \sim \mathcal{N}_p(\theta, \sigma^2I_p)$ and $Y \sim \mathrm{SG}(2\varepsilon^2)$.
Our goal is to compare the risks (expectation of the loss) of the naive estimator $\hat{\theta}^{0} = Z$ and the JS-estimator\[
\hat{\theta}^{\mathrm{JS}} = \left(1 - \frac{(p - 2)\hat{\sigma}^2}{\|Z\|^2_2} \right) Z.
\]
under squared error loss. It is well known that $\hat{\theta}^{\mathrm{JS}}$
strictly dominates $\hat{\theta}^{0}$ when $Y=0$ and $p \geq 3$ \citep{James-Stein61}. We aim to show that this dominance remains valid even under additive sub-Gaussian noise, provided the proxy variance $2\varepsilon^2$ is sufficiently small.

We first observe that it suffices to consider the case $\sigma^2=1$ The result for general $\sigma^2$ then follows by a simple scaling argument. Our {\it first key result} is the following dominance theorem for the Gaussian random variable:
\begin{tcolorbox}[colback=gray!5!white,colframe=gray!75!black,title=JS-Estimator Dominance under Sub-Gaussian Perturbations for the Gaussian Model]
\begin{theorem} \label{Dom-Gaus}
Let $\theta \in \mathbb{R}^p$ be an unknown parameter vector. Suppose that we observe $
Z = X + Y$,
where $X \sim \mathcal{N}_p(\theta, I_p)$ and $Y \in \mathbb{R}^p$ is an independent $Y \sim \mathrm{SG}(2\varepsilon^2)$.
Define the JS-estimator by
\[
\hat{\theta}^{\mathrm{JS}} = \left(1 - \frac{p - 2}{\|Z\|^2} \right) Z,
\]
and the observation-based estimator by $\hat{\theta}^{0} = Z$.

Then, for all $\theta \in \mathbb{R}^p$ and for all $p \geq 3$ the JS-estimator strictly dominates the observation-based estimator in mean squared risk:
\[
R(\hat{\theta}^{\mathrm{JS}}, \theta) < R(\hat{\theta}^{\mathrm{0}}, \theta).
\]
\end{theorem}
\end{tcolorbox}

We now provide a sketch of the proof for this theorem and refer the reader to the Appendix \ref{Appendix} for a detailed proof.

\begin{proof}[Sketch of the Proof]
We can rewrite the risk of the JS-estimator as: 
\begin{equation}
\begin{aligned}
R(\hat{\theta}^{\mathrm{JS}}, \theta) = R(\hat{\theta}^{\mathrm{0}}) - 2(p-2) \mathbb{E}\left(\frac{Z^{\top}(Z-\theta)}{Z^{\top} Z}\right)+(p-2)^{2} \mathbb{E}\left(\frac{1}{Z^{\top} Z}\right).
\end{aligned}
\label{JS-risk1}
\end{equation}
The last two terms in equation \ref{JS-risk1} can be rewritten as:
$$
\begin{aligned}
-2(p-2) \mathbb{E}\left(\frac{Z^{\top}(Z-\theta)}{Z^{\top} Z}\right)+(p-2)^{2} \mathbb{E}\left(\frac{1}{Z^{\top} Z}\right) = (p-2)\left(-2+\mathbb{E}\left(\frac{2 Z^{\top} \theta+p-2}{Z^{\top} Z}\right)\right)
\end{aligned}
$$
Further, by applying Cauchy–Schwarz and Jensen's inequalities we prove (in the Appendix) that:
$$\mathbb{E}\left(\frac{2 Z^{\top} \theta+p-2}{Z^{\top} Z}\right)<2.$$
Hence, the risk of the JS-estimator is strictly less than that of the observation-based estimator for $p \geq 3$, thus completing the proof.
\end{proof}
We are now ready to address the issue of shrinkage estimation of the variance parameter in the above normally distributed random variable with additive sub-Gaussian perturbation. 

\subsection{Dominance of the JS-estimator under Sub-Gaussian Perturbations for Gamma Distributed Random Variables}

In many applications, SVs follow a scaled chi-squared distribution, which is a special case of the gamma distribution. Given observations \( x_{ij} \), define the SV:
$\hat{\sigma}_i^2 = \frac{1}{n} \sum_{j=1}^n (x_{ij} - \bar{x}_i)^2 \sim \mathrm{Gamma}\left(\alpha = \frac{n-1}{2},\ \beta_i = \frac{2\sigma_i^2}{n}\right)$.
Thus, estimating \( \sigma_i^2 \) reduces to estimating the scale parameter \( \beta_i \) in a Gamma distribution with fixed shape parameter \( \alpha \). For this task, we can use the following theorem from literature \citep{DasGupta86AOS}:

\begin{theorem}[Dominance of the JS-Estimator for the Gamma Model]\mbox{}\\
Let $X_1, \dots, X_p \sim \mathrm{Gamma}(\alpha, \beta_i)$ be independently distributed. Consider the naive estimator
$\hat{\beta}_i^{\,0} = \frac{X_i}{\alpha + 1}, i=1,\dots,p.$
Define the JS-type shrinkage estimator
$ \hat{\beta}_i^{\,\mathrm{JS}} = \hat{\beta}_i^{\,0} + c V, V = \Biggl(\prod_{j=1}^p X_j \Biggr)^{\!1/p}.$
If $c \in \left[\,0, \; \frac{2(p-1)}{(\alpha+1)(\alpha p+1)} \,\right],$
then the shrinkage estimator dominates the naive estimator in risk, i.e. $R\!\left(\hat{\beta}^{\,\mathrm{JS}}, \beta\right) 
\;\leq\; R\!\left(\hat{\beta}^{\,0}, \beta\right), \forall \; \beta \in \mathbb{R}_+^p.$
\end{theorem}

Applying the previous theorem to SVs, we obtain the JS-type shrinkage estimator
\[
\hat{\sigma}_i^{2,\,\mathrm{JS}} 
\;=\; \frac{n}{n+1}\,\hat{\sigma}_i^2 \;+\; c\,V, 
\qquad 
V \;=\; \Biggl(\prod_{j=1}^p \hat{\sigma}_j^2 \Biggr)^{\!1/p}, 
\qquad
c \;\in\; \left[\,0, \; \frac{4 n (p-1)}{(n+1)\big((n-1)p+2\big)} \,\right].
\]

We now present our \emph{second key result}, namely, that this dominance persists even under additive sub-Gaussian noise for the Gamma-distributed random variables.

\begin{tcolorbox}[colback=gray!5!white,colframe=gray!75!black,title=JS-Estimator Dominance under Sub-Gaussian Perturbations for the Gamma Model]
\begin{theorem}
\label{Dom-Gamma}
Let \( Z_{ij} = X_{ij} + Y_{ij} \), where \( X_{ij} \sim \mathcal{N}(\mu, \sigma_{x,i}^2) \), \( Y_{ij} \sim \mathrm{SG}(2\varepsilon^2) \), and define
$
\hat{\sigma}_{z,i}^2 = \hat{\sigma}_{x,i}^2 + \hat{\sigma}_{y,i}^2 + W_i,$
where \( W_i = \frac{2}{n} \sum_j (X_{ij} - \bar{X}_i)(Y_{ij} - \bar{Y}_i) \). Then, the estimators
\[
\hat{\beta}_i^{\,0} = \frac{\hat{\sigma}_{z,i}^2}{\alpha + 1}, \quad
\hat{\beta}_i^{\,\mathrm{JS}} = \hat{\beta}_i^{\,0} + cV, \quad
V = \left( \prod_{j=1}^p \hat{\sigma}_{z,j}^2 \right)^{1/p}
\]
satisfy
\[R(\hat{\beta}^{\,\mathrm{JS}}, \beta) < R(\hat{\beta}^{\,0}, \beta)\]
if
\[
0 < c < \frac{2p}{\alpha p + 1} \exp\left(\frac{1}{\alpha}\right) \sqrt{1 + \frac{1}{\alpha}} - \frac{2}{\alpha + 1}, \quad \text{or} \quad c \leq -\frac{2}{\alpha(\alpha + 1)}.
\]
\end{theorem}
\end{tcolorbox}
\begin{remark}
The admissible range for \( c \) under sub-Gaussian perturbations includes the classical interval, so standard shrinkage remains valid even in noisy settings.
\end{remark}

\begin{proof}[Sketch of the Proof]
We aim to prove that the JS-type estimator $\hat{\beta}^{\,\mathrm{JS}}$ has strictly smaller risk than the naive estimator $\hat{\beta}^{\,0}$, i.e.
$R(\hat{\beta}^{\,\mathrm{JS}}, \beta) < R(\hat{\beta}^{\,0}, \beta)$. For a detailed proof, the reader is referred to the Appendix \ref{Appendix}.

We begin by expressing the risk of $\hat{\beta}^{\,\mathrm{JS}}$  as:
$$
\begin{aligned}
R(\hat{\beta}^{\,\mathrm{JS}}, \beta) =\mathbb{E}\left(\left\|\hat{\beta}^{\,\mathrm{JS}}-\beta\right\|_{2}^{2}\right)=\mathbb{E}\left(\left\|\hat{\beta}^{\,0}+c V-\beta\right\|_{2}^{2}\right)=\\
=\mathbb{E}\left(\left\|\hat{\beta}^{\,0}-\beta\right\|_{2}^{2}\right)+2 \sum_{i=1}^{p} \mathbb{E}\left(\left(\hat{\beta}_{0, i}-\beta_{i}\right) \cdot c V\right)+p \mathbb{E}\left(c^{2} V^{2}\right)
\end{aligned}
$$
Thus, the difference in risks depends on the sign of the second term in the above expression expressed as:
$$c \left( 2 \sum_{i=1}^{p} \mathbb{E}\left(\left(\hat{\beta}_{0, i}-\beta_{i}\right) V \right)+p \mathbb{E}\left(c V^{2}\right) \right).$$

To show that this quantity is negative, we analyze the sign depending on whether $c>0$ or $c<0$, and whether the linear term dominates the quadratic term. We consider two main cases:
\begin{itemize}
    \item[\textbf{Case 1:}] \( c < 0 \) and 
    \begin{equation}
        2 \sum_{i=1}^{D} \mathbb{E}\left[\left(\hat{\beta}_{0,i} - \beta_i\right)V\right] 
        + p c\, \mathbb{E}\left[V^2\right] > 0
        \label{eq:case1_condition}
    \end{equation}
    
    \item[\textbf{Case 2:}] \( c > 0 \) and 
    \begin{equation}
        2 \sum_{i=1}^{p} \mathbb{E}\left[\left(\hat{\beta}_{0,i} - \beta_i\right)V\right] 
        + p c\, \mathbb{E}\left[V^2\right] < 0
        \label{eq:case2_condition}
    \end{equation}
\end{itemize}
In both cases outlined above, the proof fundamentally relies on an application of Stein's lemma for the Gamma distribution given below. Since each \( \hat{\sigma}^2_{x,i} \sim \mathrm{Gamma}(\alpha, \beta) \) with known shape and rate parameters, we can invoke the Stein's identity given in the following lemma:

\begin{lemma}[Stein's Lemma for the Gamma Distribution]
Let \( X \sim \mathrm{Gamma}(\alpha, \beta) \), and let \( h(x) \) be a differentiable function such that 
$$\mathbb{E}[X h(X)] < \infty \quad \text{and} \quad \mathbb{E}[X h'(X)] < \infty.$$
Then the following (Stein's) identity holds:
$$\mathbb{E}[(X - \alpha \beta) h(X)] = \beta\, \mathbb{E}[X h'(X)].$$
\end{lemma}

We apply Stein's identity to the function \( h(x) = V \), where $V$ is the geometric mean defined by
$
V = \left[ \prod_{i=1}^p \left( \hat{\sigma}_{x,i}^2 + \hat{\sigma}_{y,i}^2 + w_i \right) \right]^{1/p}, \quad X_i = \hat{\sigma}_{x,i}^2
$.
Using the Stein's identity for the Gamma distributed random variable, which is given by,
$$\mathbb{E}[h(X)] = \frac{1}{\alpha \beta} \mathbb{E}[X h(X)] - \frac{1}{\alpha} \mathbb{E}[X h'(X)],$$
we differentiate \( h(X) \) with respect to \( \hat{\sigma}_{x,k}^2 \) and get:
$
\frac{dV}{d \hat{\sigma}_{x,k}^2} = \frac{1}{p} \cdot V \cdot \left(\frac{1}{\hat{\sigma}_{x,k}^2 + \hat{\sigma}_{y,k}^2 + w_k}\right).
$
Substituting into Stein's identity for \( X_i = \hat{\sigma}_{x,i}^2 \), we get:
$$\mathbb{E}[V] = \frac{1}{\alpha \beta_i} \mathbb{E}\left[ \hat{\sigma}_{x,i}^2 \cdot V \right] - \frac{1}{\alpha p} \mathbb{E} \left[ V \cdot \frac{ \hat{\sigma}_{x,i}^2 }{ \hat{\sigma}_{x,i}^2 + \hat{\sigma}_{y,i}^2 + w_i } \right].$$

Depending on the sign and structure of the risk difference (as in the two main cases above), we proceed to estimate the relevant expectations using various inequalities — such as Bernstein's inequality (to control sub-exponential terms like $W_i$), Jensen's inequality, Cauchy–Schwarz, and the arithmetic–geometric mean inequality respectively. These tools allow us to bound the expressions for a given choice of $c$ and derive the dominance for the JS-estimator of the variance parameter of the normally distributed random variable under sub-Gaussian perturbations.
Further technical details and precise bounds are deferred to the appendix.
\end{proof}

%% file: experiments.tex
\section{Experiments}\label{expts}
In this section, we present several experimental results demonstrating the performance of BN with our proposed shrinkage estimators in both image classification and segmentation tasks, under standard and adversarial settings. To simulate adversarial attacks, we generated samples of the noise signal modeled by a sub-Gaussian–like distribution with the density $$f(x) = \frac{1}{2\sqrt{2}\pi\sigma}\left(\frac{x^2}{2 \sigma^2} + \frac{1}{4}\right)^{-1},$$ which arises from a mixture of Lévy and Gaussian distributions \citep{GEORGIOU20063061}. The samples were drawn using inverse transform sampling.

For classification, we employed the ResNet-9 \citep{he2015deepresiduallearningimage} architecture on the CIFAR-10  dataset \citep{cifar10}, as well as a 3D CNN model on the PPMI dataset \citep{ppmi-data}. For object segmentation, we used the HRNetV2 \citep{wang2020deep} model applied to the Cityscapes dataset \citep{Cordts2015Cvprw}. Since BN is known to become unstable when the batch size is small, we report results for varying batch sizes and sub-Gaussian noise levels. Our goal is to demonstrate that our shrinkage-based method is more robust to adversarial perturbations and offers greater overall stability.

{\it We remind the reader that our goal here is not to improve the SOTA results on these data in the literature, but to demonstrate the performance gains from the use of BN in the presence (absence) of adversarial attacks with the correct use of Stein shrinkage compared to vanilla BN and other forms of shrinkage applied to BN. Throughout all experiments, we avoided training-time tricks and focused on obtaining an honest comparison.}

In the experiments, we compare our approach mainly with \cite{khoshsirat2023improvingnormalizationjamessteinestimator}, aiming to show that theoretically correct estimation of mean as well as variance shrinkage significantly improves performance, particularly in small batch regimes. In addition, with the correct variance shrinkage, one can achieve higher accuracy in the presence of adversarial perturbations.
We also tested a variant using only the mean shrinkage estimator and no shrinkage on variance. Additionally, we included models regularized with Lasso and Ridge methods for comparison, as these are classical shrinkage methods in literature. Below we summarize the forms of these shrinkage estimators  and provide explicit formulas for convenience. Derivations can be found, for example, in \cite{hastie2009elements}.

\textbf{Lasso mean and variance shrinkage.} 
These estimators are obtained by applying an $\ell_1$ penalty in penalized least squares problems (\cite{hastie2009elements}). 
For the mean, this leads to the soft-thresholding solution
$\hat{\mu} = \operatorname{sign}(\bar{x}) \max\!\left(0, |\bar{x}| - \tfrac{\lambda}{2n}\right),$
where $\bar{x} = \tfrac{1}{n} \sum_{i=1}^n x_i$ is the SM. 
This form encourages the shrinkage of small mean values towards zero, potentially setting them exactly to zero when $\lambda$ is large. 
For variance, an analogous penalized formulation yields
$\hat{\sigma}^2 = \max\!\left(0,\, s^2 - \tfrac{\lambda}{2}\right),$
where $\lambda > 0$ is the regularization hyperparameter and
$s^2 = \tfrac{1}{n}\sum_{i=1}^n(x_i - \bar{x})^2$ is the SV.

\textbf{Ridge mean and variance shrinkage.} 
These estimators are obtained by applying an $\ell_2$ penalty in penalized least squares problems (\cite{hoerl1970ridge, hastie2009elements}). 
For the mean, this yields
$\hat{\mu} = \frac{\sum_{i=1}^n x_i}{n + \lambda},$
and for the variance
$\hat{\sigma}^2 = \frac{s^2}{1 + \lambda},$
where $s^2 = \tfrac{1}{n}\sum_{i=1}^n (x_i - \bar{x})^2$ is the SV 
and $\lambda > 0$ is the regularization parameter.

Complete experimental setup are provided in the Appendix B. Note that in all the tables depicting the results, we have highlighted the best performing BN method (ours) for small batch sizes that are preferred from a computational efficiency perspective as well as when data are scarce. Code available at {\href{https://github.com/sivolgina/stein_shrinkage}{https://github.com/sivolgina/shrinkage}.

\subsection{Application to Cifar10 Data}
We conducted experiments on the CIFAR-10 dataset, which consists of 50,000 training and 10,000 test images across 10 classes. We simulate sub-gaussian noise by perturbing the input features with zero-mean noise of varying magnitudes $(0-30\%)$, applied independently to each image during validation and testing. The results in Table~\ref{tab:noise_levels_comparison_cifar} confirm that our proposed BN significantly outperforms the baselines in noisy settings across all batch sizes. Although standard BN variants quickly degrade under noise, our method maintains substantially higher accuracy, especially with moderate to large noise levels $(10-30\%).$ Compared to the method in \cite{khoshsirat2023improvingnormalizationjamessteinestimator}, our approach shows consistent improvement due to the correct development and use of variance shrinkage. The method with only mean correction performs worse than our full model, highlighting the importance of variance correction. Finally, Lasso and Ridge-based approaches do not improve noise performance, showing behavior similar to the standard BN.

\begin{table}[t!]
\centering
\resizebox{\textwidth}{!}{
\begin{tabular}{lcccccc}
\toprule
\textbf{Method} & \textbf{0\% noise} & \textbf{3\% noise} & \textbf{10\% noise} & \textbf{15\% noise} & \textbf{20\% noise} & \textbf{30\% noise}\\
\midrule
Standard (32) & 79.2 $\pm$ 0.28 & 29.83 $\pm$ 2.51 & 17.19 $\pm$ 1.6 & 15.28 $\pm$ 1.38 & 14.38 $\pm$ 0.83 & 14.22 $\pm$ 1.32 \\
Standard (64) & 81.7 $\pm$ 0.27 & 30.17 $\pm$ 2.73 & 19.19 $\pm$ 2.05 & 17.10 $\pm$ 1.54 & 15.54 $\pm$ 2.07 & 13.7 $\pm$ 0.93 \\
Standard (128) & 79.8 $\pm$ 0.3 & 29.05 $\pm$ 3.63 & 19.48 $\pm$ 2.11 & 16.23 $\pm$ 1.36 & 19.05 $\pm$ 3.63 & 19.48 $\pm$ 2.11 \\
\midrule
Ours (32) & {\bf 80.89} $\pm$ 0.51 & {\bf 60.7} $\pm$ 0.96 & {\bf 44.55} $\pm$ 1.4  & {\bf 35.34} $\pm$ 1.6  & {\bf 26.67} $\pm$ 3.38  & {\bf 25.66} $\pm$ 3.63\\
Ours (64) & 80.78 $\pm$ 0.54 & 61.7 $\pm$ 0.81 & 45.17 $\pm$ 1.43  & 36.19 $\pm$ 0.57  & 32.05 $\pm$ 0.58  & 25.26 $\pm$ 0.69\\
Ours (128) & 79.1 $\pm$ 0.3 & 60.49 $\pm$ 1.06 & 44.96 $\pm$ 1.3  & 36.27 $\pm$ 1.01  & 29.2 $\pm$ 4.67  & 28.12 $\pm$ 3.32\\
\midrule
Mean corrected (32) & 78.7 $\pm$ 1.32 & 56.79 $\pm$ 0.87 & 40.13 $\pm$ 1.19 & 31.57 $\pm$ 1.19 & 25.82 $\pm$ 3.65 & 24.77 $\pm$ 3.97 \\
Mean corrected (64) & 80.6 $\pm$ 0.4 & 54.15 $\pm$ 1.26 & 38.24 $\pm$ 0.66 & 30.17 $\pm$ 1.37 & 26.78 $\pm$ 1.42 & 20.85 $\pm$ 1.13 \\
Mean corrected (128) & 79.9 $\pm$ 0.48 & 49.2 $\pm$ 1.35 & 32.96 $\pm$ 0.99 & 25.97 $\pm$ 1.2 & 25.82 $\pm$ 3.65 & 24.77 $\pm$ 3.97 \\
\midrule
Khoshsirat et al. (32) & 76.82 $\pm$ 1.01 & 57. 49 $\pm$ 1.71 & 41.93 $\pm$ 1.46  & 33.20 $\pm$ 1.73 & 27.10 $\pm$ 3.5  & 25.39 $\pm$ 3.72\\
Khoshsirat et al. (64) & 79.84 $\pm$ 1.18 & 58.71 $\pm$ 1.48 & 43.75 $\pm$ 1.67  & 33.92 $\pm$ 2.4 & 30.75 $\pm$ 1.32  & 23.36 $\pm$ 1.20 \\
Khoshsirat et al. (128) & 79.12 $\pm$ 0.27 & 60.02 $\pm$ 1.11 & 44.85 $\pm$ 1.47  & 36.35 $\pm$ 1.47 & 29.15 $\pm$ 4.25  & 28.34 $\pm$ 3.26 \\
\midrule
Lasso (32) & 78.4 $\pm$ 1.0 & 27.50 $\pm$ 3.88 & 16.43 $\pm$ 2.42 & 14.79 $\pm$ 2.08 & 14.9 $\pm$ 1.59 & 14.28 $\pm$ 1.32 \\
Lasso (64) & 80.3 $\pm$ 0.6 & 28.42 $\pm$ 3.75 & 17.54 $\pm$ 3.29 & 16.47 $\pm$ 1.44 & 16.27 $\pm$ 0.9 & 13.91 $\pm$ 1.12 \\
Lasso (128) & 79.3 $\pm$ 0.38 & 25.79 $\pm$ 6.26 & 17.77 $\pm$ 3.04 & 15.22 $\pm$ 1.45 & 14.27 $\pm$ 1.91 & 13.96 $\pm$ 1.93 \\
\midrule
Ridge (32) & 78.4 $\pm$ 0.69 & 30.13 $\pm$ 4.58 & 17.49 $\pm$ 2.89 & 15.29 $\pm$ 2.18 & 14.68 $\pm$ 2.13 & 14.37 $\pm$ 0.69 \\
Ridge (64) & 80.6 $\pm$ 0.41 & 27.36 $\pm$ 3.8 & 18.32 $\pm$ 3.09 & 15.0 $\pm$ 1.22 & 15.62 $\pm$ 0.56 & 13.4 $\pm$ 0.61 \\
Ridge (128) & 79.9 $\pm$ 0.48 & 28.36 $\pm$ 3.24 & 19.53 $\pm$ 1.82 & 16.67 $\pm$ 1.4 & 13.48 $\pm$ 1.64 & 13.77 $\pm$ 2.18 \\
\bottomrule
\end{tabular} }
\caption{Performance of ResNet-9 with different BN methods under varying subgaussian noise and batch sizes respectively. Values depict mean accuracy $\pm$ standard deviation on test CIFAR-10 data. Numbers inside parentheses in the far left column denote batch size.
}
\label{tab:noise_levels_comparison_cifar} 
\end{table}

\subsection{Application to Cityscapes Data}

\begin{table}[t!]
\centering
\resizebox{\textwidth}{!}{
\begin{tabular}{lcccccc}
\toprule
\textbf{Method} & \textbf{0\% noise} & \textbf{3\% noise} & \textbf{10\% noise} & \textbf{15\% noise} & \textbf{20\% noise} & \textbf{30\% noise}\\
\midrule
Standard (8) & 68.1 $\pm$ 0.9  & 8.6 $\pm$ 0.8 & 7 $\pm$ 0.9 & 6.14 $\pm$ 1.6 & 5.7 $\pm$ 2.0 & 4.8 $\pm$ 2.9 \\
Standard (16) & 67.4 $\pm$ 0.9 & 9.9 $\pm$ 1.7 & 7.6 $\pm$ 2.1 & 6.8 $\pm$ 2.4 & 6.4 $\pm$ 2.5 & 5.7 $\pm$ 2.6 \\
\midrule
Ours (8) & {\bf 68.8} $\pm$ 0.1 & {\bf 39.6} $\pm$ 2.3 & {\bf 32.5} $\pm$ 2.6  & {\bf 29.9} $\pm$ 2.4  & {\bf 28.9} $\pm$ 2.3  & {\bf 27.2} $\pm$ 1.8\\
Ours (16) & 68.4 $\pm$ 0.4 & 36.3 $\pm$ 2.7 & 29.4 $\pm$ 1.5  & 27.1 $\pm$ 1.5  & 26.3 $\pm$ 1.7  & 25.1 $\pm$ 2.2\\
\midrule
Mean corrected (8) &  67.80 $\pm$ 0.2 & 39.13 $\pm$ 2.6 & 31.8 $\pm$ 2.1 & 29.1 $\pm$ 2 & 28.1 $\pm$ 2 & 26.4 $\pm$ 2.2 \\
Mean corrected (16) & 68.1 $\pm$ 0.5 & 39.07 $\pm$ 1.7 & 32.8 $\pm$ 1 & 30.8 $\pm$ 1.2 & 30 $\pm$ 1.3 & 28.6 $\pm$ 1.2 \\
\midrule
Khoshsirat et al. (8) & 68.4 $\pm$ 0.3 & 36.4 $\pm$ 2.8 & 28.5 $\pm$ 3.5  & 25.9 $\pm$ 3.5 & 24.8 $\pm$ 3.2  & 23.3 $\pm$ 3 \\
Khoshsirat et al. (16) & 67.1 $\pm$ 0.3 & 36.2 $\pm$ 0.7 & 28.7 $\pm$ 1.3  &  26.3 $\pm$ 1.3 & 25.5 $\pm$ 1.2  & 24.4 $\pm$ 1.2 \\
\midrule
Lasso (8) & 65.9 $\pm$ 0.2 & 9.7 $\pm$ 4.2 & 6.2 $\pm$ 2.5 & 5.1 $\pm$ 2.2 & 4.6 $\pm$ 2.1 & 3.9 $\pm$ 1.9 \\
Lasso (16) & 64.9 $\pm$ 0.9 & 8 $\pm$ 4.3 & 5.5 $\pm$ 3.2 & 4.7 $\pm$ 3.1 & 4.3 $\pm$ 3.1 & 3.7 $\pm$ 3 \\
\midrule
Ridge (8) & 68.4 $\pm$ 0.9 & 8.4 $\pm$ 1.4 & 7.3 $\pm$ 1.6 & 7.1 $\pm$ 1.9 & 7 $\pm$ 2.1 & 6 $\pm$ 2.3 \\
Ridge (16) & 66.8 $\pm$ 0.1 & 6.3 $\pm$ 1.9 & 4.6 $\pm$ 2.1 & 4.2 $\pm$ 2.5 & 4.1 $\pm$ 2.6 & 3.9 $\pm$ 2.7 \\
\bottomrule
\end{tabular} }
\caption{Performance of HRNetV2 with different BN  methods under varying subgaussian noise and batch sizes respectively. Values depict mean of mIoU scores $\pm$ standard deviation on part of validation Cityscapes data using as a test data. Numbers inside paranthesis in the far left column denote batch size.}
\label{tab:noise_levels_comparison_cityscapes} 
\end{table}

To further evaluate the robustness of different BN variants, we tested them on the Cityscapes dataset -- a more challenging task that involves high resolution urban scene segmentation with fine-grained object boundaries. The dataset contains 5,000 finely annotated images from 50 cities with 19 semantic classes used for evaluation.
We use the mean intersection over union (mIoU), a standard metric used for semantic segmentation in the computer vision community.
As in the previous experiment, we injected sub-gaussian noise into the feature maps during validation and testing to simulate corrupted statistics. This setup is particularly relevant for segmentation, where maintaining accurate local feature distributions is crucial for pixel-level predictions.
As shown in Table~\ref{tab:noise_levels_comparison_cityscapes}, standard BN fails to retain meaningful performance even under mild noise (e.g., below $10$ mIoU score at $3\%$ corruption). In contrast, our proposed method and its mean corrected variant significantly improve robustness, maintaining mIoU scores above $25\%$ even at high noise settings. Regularization-based approaches (Lasso and Ridge) show negligible improvement under noise and have a behavior similar to that of standard BN. 

\subsection{Application to PPMI Data}
In this experiment, the goal is to demonstrate the performance of our BN method on 3D diffusion tensor magnetic resonance images (DT-MRI or in short DTI) acquired from Parkinson's Disease (PD) affected brains and normal brains (Controls).  DTI is a noninvasive MRI technique that allows one to infer neuronal connectivity patterns in the tissue being imaged. The PPMI data were subjected to a standard preprocessing pipeline (see Appendix B).
using the FSL software \citep{FSLpack}. DTIs are matrix-valued images where each voxel has a $(3,3)$ symmetric positive definite matrix that captures the covaraince of a local (voxel-level) Gaussian fit.
These DTIs are further processed to extract scalar-valued feature maps, called fractional anisotropy (FA) maps. The FA is a scalar function of the three eigen values of the diffusion tensor at each voxel in the DTI. Harmonization \citep{FORTIN2017149} is applied to these FA maps to remove site effects caused by distinct acquisition protocols and equipment used. 
The data set is highly imbalanced and contains FA maps from 360 Control and 130 PD brain scans. This imbalance is addressed using mixup-based augmentation \citep{DBLP:journals/corr/abs-1710-09412} with sampling bias favoring the smaller sized class (PD). Using the same setup as in previous experiments and a 5-fold cross-validation to compare 3D CNN with different variants of 3D BN, we get results summarized in the Table \ref{tab:noise_levels_comparison_fa}. These results further support our earlier observations, even in this  challenging classification task. Since the image data in this experiment are 3D volumetric images, the computational needs are much higher than when handling 2D image data sets, and we therefore used smaller batch sizes in the reported results. This helps reduce the computational footprint of the experiment. 

\begin{table}[t!]
\centering
\resizebox{\textwidth}{!}{
\begin{tabular}{lcccccc}
\toprule
\textbf{Method} & \textbf{0\% noise} & \textbf{3\% noise} & \textbf{10\% noise} & \textbf{15\% noise} & \textbf{20\% noise} & \textbf{30\% noise}\\
\midrule
Standard (10) & 94.9 $\pm$ 1.3 & 71.2 $\pm$ 1.6 & 64.7 $\pm$ 4.3 & 62.3 $\pm$ 5 & 61.3 $\pm$ 3.9 & 60.7 $\pm$ 8.2 \\
Standard (16) & 93.3 $\pm$ 1.5 & 72.6 $\pm$ 1.5 & 63.7 $\pm$ 6.7 & 63.5 $\pm$ 4.6 & 62.3 $\pm$ 5.6 & 59.1 $\pm$ 5.4 \\
\midrule
Ours (10) &  95.1 $\pm$ 1.7 & {\bf 71.8} $\pm$ 2.8 & {\bf 68} $\pm$ 3.6  & {\bf 68} $\pm$ 3.9  & {\bf 66.8} $\pm$ 5.2  & {\bf 66.5} $\pm$ 6.9\\
Ours (16) & 95.1 $\pm$ 1.8 & 72.4 $\pm$ 2.8 & 65.7 $\pm$ 3.6  & 63.5 $\pm$ 6  & 61.5 $\pm$ 5.3  & 60.9 $\pm$ 7.2\\
\midrule
Mean corrected (10)  & 95.3 $\pm$ 2.6 & 68.6 $\pm$ 7.3 & 64.7 $\pm$ 6.1 & 64.1 $\pm$ 5.7 & 63.7 $\pm$ 6.1 & 62.3 $\pm$ 4.3 \\
Mean corrected (16)  & 93.1 $\pm$ 2.4 & 69.2 $\pm$ 7.9 & 62.9 $\pm$ 12.2 & 61.1 $\pm$ 7.5 & 59.7 $\pm$ 10.9 & 58.7 $\pm$ 8.2 \\
\midrule
Khoshsirat et al. (10)  & 95.1 $\pm$ 2.4 & 71.4 $\pm$ 2.4 & 67.6 $\pm$ 4.5  & 66.4 $\pm$ 4.5 & 65.3 $\pm$ 7  & 64.9 $\pm$ 7.2 \\
Khoshsirat et al. (16)  & 94.7 $\pm$ 2.5 & 70.8 $\pm$ 1.9 & 67.2 $\pm$ 3.9  &  65.1 $\pm$ 6.1 & 64.3 $\pm$ 3.7  & 63.5 $\pm$ 4.6 \\
\midrule
Lasso (10)  & 96.5 $\pm$ 3.0 & 64.9 $\pm$ 6.9 & 62.5 $\pm$ 9.4 & 62.3 $\pm$ 12.9 & 60.7 $\pm$ 9.3 & 57.5 $\pm$ 7.1 \\
Lasso (16)  & 95.1 $\pm$ 1.9 & 65.3 $\pm$ 3.9 & 64.5 $\pm$ 5.2 & 62.9 $\pm$ 4.9 & 59.1 $\pm$ 4.6 & 57.2 $\pm$ 4.4 \\
\midrule
Ridge (10) & 95.7 $\pm$ 1.7 & 68.8 $\pm$ 3.6 & 64.7 $\pm$ 4.1 & 64.5 $\pm$ 6.5 & 62.9 $\pm$ 4 & 61.7 $\pm$ 5.3 \\
Ridge (16) & 97.1 $\pm$ 1.3 & 71 $\pm$ 1.8 & 64.3 $\pm$ 7.4 & 63.7 $\pm$ 3.3 & 62.1 $\pm$ 5.5 & 61.1 $\pm$ 2.6 \\
\bottomrule
\end{tabular} }
\caption{3D CNN performance with different BN  methods under varying subgaussian noise and batch sizes respectively on FA maps data. Values depict mean of accuracies $\pm$ standard deviation. Numbers inside paranthesis in the far left column denote batch size.}
\label{tab:noise_levels_comparison_fa} 
\end{table}

\subsection{Experiments using Data Dependent Attacks}

To validate our theoretical findings, we conducted a series of experiments using well-known adversarial attacks. While numerous attack methods have been proposed, our focus here is on confirming the theoretical insights rather than providing an extensive empirical comparison. Therefore, we restricted our study to two of the most widely used attacks, FGSM \citep{goodfellow2015explaining} and it's iterative version PGD. The results are presented below; for detailed experimental setups, we refer the reader to the appendix.

We first evaluate robustness against the FGSM attack. As before, we use a ResNet-9 model trained on the CIFAR-10 dataset and HRNetV2 model on the Cityscapes dataset, with the same hyperparameters as in the previous experiments. For consistency, we consider batch sizes matching those used in the earlier experiments and run each configuration five times. The results at different perturbation levels ($\epsilon = 0.03, 0.1, 0.15, 0.2, 0.3$) are reported in Tables \ref{tab:fgsm1}, \ref{tab:fgsm2}. We omit regularized BN baselines on CIFAR-10, as they perform similarly to standard BN.

\begin{table}[t!]
\centering
\scalebox{0.925}{
\begin{tabular}{lcccccc}
\toprule
\textbf{BN Method} & \textbf{$\epsilon = 0.03$} & \textbf{$\epsilon = 0.1$} & \textbf{$\epsilon = 0.15$} & \textbf{$\epsilon = 0.2$} & \textbf{$\epsilon = 0.3$}\\
\midrule
Standard (32) & 0.26 $\pm$ 0.04 & 0.07 $\pm$ 0.05 & 0.18 $\pm$ 0.18 & 0.45 $\pm$ 0.53 & 1.0 $\pm$ 0.96 \\
Standard (64) & 0.1 $\pm$ 0.05 & 0.01 $\pm$ 0.05 & 0.04 $\pm$ 0.02 & 0.13 $\pm$0.08 & 0.32$\pm$ 0.11 \\
Standard (128) & 0.05 $\pm$ 0.02 & 0.06 $\pm$ 0.05 & 0.05 $\pm$ 0.02 & 0.11 $\pm$ 0.04 & 0.39 $\pm$ 0.13 \\
\midrule
Ours (32) & {\bf 42.72} $\pm$ 1.16 & {\bf 26.06} $\pm$ 1.5  & {\bf 21.66} $\pm$ 1.31  & {\bf 18.74} $\pm$ 0.99  & {\bf 14.868} $\pm$ 0.94\\
Ours (64) & {\bf 43.29} $\pm$ 0.56 & {\bf 26.17} $\pm$ 0.41  & {\bf 21.66} $\pm$ 0.34 & {\bf 18.69} $\pm$ 0.39  & {\bf 14.8} $\pm$ 0.35\\
Ours (128) & {\bf 36.36} $\pm$ 1.38 & {\bf 20.27} $\pm$ 1.35  & {\bf 16.82} $\pm$ 1.27  & {\bf 14.41} $\pm$ 1.16  & {\bf 11.45} $\pm$ 0.97\\
\midrule
Mean corrected (32) & 34.39 $\pm$ 0.84 & 20.07 $\pm$ 0.72 & 16.53 $\pm$ 0.76 & 14.28 $\pm$ 0.6 & 11.42 $\pm$ 0.49 \\
Mean corrected (64) & 26.82 $\pm$ 1.08 & 14.4 $\pm$ 1.16 & 11.99 $\pm$ 0.99 & 10.29 $\pm$ 0.84 & 7.98 $\pm$ 0.59 \\
Mean corrected (128) & 16.57 $\pm$ 0.96 & 8.43 $\pm$ 1.19 & 7.03 $\pm$ 1.17 & 6.14 $\pm$ 1.0 & 4.92 $\pm$ 0.83 \\
\midrule
Khoshsirat et al. (32) & 37.04 $\pm$ 0.71 & 22.78 $\pm$ 0.92  & 19.06 $\pm$ 0.86 & 16.35 $\pm$ 0.66  & 12.75 $\pm$ 0.49 \\
Khoshsirat et al. (64) & 38.27 $\pm$ 1.01 & 22.97 $\pm$ 0.49  & 19.07 $\pm$ 0.62 & 16.24 $\pm$ 0.73  & 12.3 $\pm$ 0.64 \\
Khoshsirat et al. (128) & 33.45 $\pm$ 0.79 & 18.49 $\pm$ 0.82  & 15.22 $\pm$ 0.71 & 12.94 $\pm$ 0.54  & 9.9 $\pm$ 0.85 \\
\bottomrule
\end{tabular} }
\caption{Performance of ResNet-9 with different BN methods under varying FGSM attack and batch sizes respectively. Values depict mean accuracy $\pm$ standard deviation on test CIFAR-10 data. Numbers inside the parentheses in the far left column denote batch size.}
\label{tab:fgsm1} 
\end{table}

\begin{table}[t!]
\centering
\scalebox{0.925}{
\begin{tabular}{lcccccc}
\toprule
\textbf{BN Method} & \textbf{$\epsilon = 0.03$} & \textbf{$\epsilon = 0.1$} & \textbf{$\epsilon = 0.15$} & \textbf{$\epsilon = 0.2$} & \textbf{$\epsilon = 0.3$}\\
\midrule
Standard (8) & 9.4 $\pm$ 1.5 & 7.6 $\pm$ 0.9 & 6.2 $\pm$ 1.7 & 4.9 $\pm$  2.3  & 3.8 $\pm$  2.6 \\
Standard (16) & 10.2 $\pm$ 1.0 & 5.5 $\pm$ 1.5 & 4.7 $\pm$ 1.7 & 4.3 $\pm$  1.9  & 3.9 $\pm$  2.0 \\
\midrule
Ours (8) & {\bf 38.1} $\pm$ 1.4 & {\bf 28.8} $\pm$ 1.8  & {\bf 25.9} $\pm$ 2.4  & {\bf 23.6} $\pm$ 3.2  & {\bf 20.9} $\pm$ 4.5 \\
Ours (16) & {\bf 35.4} $\pm$ 0.6 & {\bf 24.9} $\pm$ 1.0 & {\bf 21.8} $\pm$ 1.3 & {\bf 19.8} $\pm$ 1.6  & {\bf 17.3} $\pm$ 1.8\\
\midrule
Mean corrected (8) & 33.7 $\pm$ 1.9 & 24.4 $\pm$ 1.4 & 22.3 $\pm$ 1.2 & 20.8 $\pm$ 1.3 & 18.9 $\pm$ 1.8 \\
Mean corrected (16) & 33.5 $\pm$ 2.7 & 24.6 $\pm$ 3.3 & 21 $\pm$ 3.1 & 19.4 $\pm$ 2.9 & 17 $\pm$ 3.3 \\
\midrule
Khoshsirat et al. (8) & 35.9 $\pm$ 2.2 & 26.2 $\pm$ 2.6  & 23.4 $\pm$ 2.6 & 21.7 $\pm$ 2.6  & 19.7 $\pm$ 2.4 \\
Khoshsirat et al. (16) & 33.4 $\pm$ 1.8 & 22.9 $\pm$ 2.3  & 20.4 $\pm$ 2.4 & 18.87 $\pm$ 2.3 & 16.9 $\pm$ 2.1 \\
\midrule
Lasso (8) & 10.5 $\pm$ 5.7 & 5.7 $\pm$ 3.4  & 5.0 $\pm$ 2.9 & 4.8 $\pm$ 2.7  & 4.8 $\pm$ 2.8\\
Lasso (16) & 6.8 $\pm$ 2.9 & 3.6 $\pm$ 1.8  & 3.3 $\pm$ 1.7 & 3.5 $\pm$ 1.7  & 4.2 $\pm$ 2.3 \\
\midrule
Ridge (8) & 9.7 $\pm$ 1.8 & 6.2 $\pm$ 0.8  & 4.8 $\pm$ 0.8 & 4.3 $\pm$ 0.7  & 4.1 $\pm$ 0.8\\
Ridge (16) & 12.9 $\pm$ 6.3 & 7.0 $\pm$ 1.0  & 5.9 $\pm$ 0.9 & 5.5 $\pm$ 0.9  & 4.9 $\pm$ 1.0 \\
\bottomrule
\end{tabular} }
\caption{Performance of HRNetV2 with different BN methods under varying FGSM attack and batch sizes respectively. Values depict mean of mIoU scores $\pm$ standard deviation on part of validation Cityscapes data using as a test data. Numbers inside the parentheses in the far left column denote batch size.}
\label{tab:fgsm2} 
\end{table}

In addition, we evaluated robustness under the PGD attack. The experimental configuration mirrors that of the FGSM case; The corresponding results are summarized in Tables \ref{tab:pgd1}, \ref{tab:pgd2}.

\begin{table}[t!]
\centering
\scalebox{0.925}{
\begin{tabular}{lcccccc}
\toprule
\textbf{BN Method} & \textbf{$\epsilon = 0.03$} & \textbf{$\epsilon = 0.1$} & \textbf{$\epsilon = 0.15$} & \textbf{$\epsilon = 0.2$} & \textbf{$\epsilon = 0.3$}\\
\midrule
Standard (32) & 0.08 $\pm$ 0.02 & 0.08 $\pm$ 0.01 & 0.0 & 0.0 & 0.0 \\
Standard (64) & 0.19 $\pm$ 0.03 &  0.0 & 0.0 & 0.0 & 0.0 \\
Standard (128) & 0.0 &  0.0 & 0.0 & 0.0 & 0.0 \\
\midrule
Ours (32) & {\bf 27.46} $\pm$ 0.74 & {\bf 17.27} $\pm$ 0.48  & {\bf 13.5} $\pm$ 0.81  & {\bf 10.74} $\pm$ 0.64  & {\bf 6.8} $\pm$ 0.47\\
Ours (64) & {\bf 32.05} $\pm$ 1.86 & {\bf 21.7} $\pm$ 1.5  & {\bf 17.01} $\pm$ 1.4 & {\bf 13.31} $\pm$ 1.18  & {\bf 8.44} $\pm$ 0.69\\
Ours (128) & {\bf 24.54} $\pm$ 1.7 & {\bf 14.86} $\pm$ 1.54  & {\bf 11.29} $\pm$ 1.39  & {\bf 8.88} $\pm$ 1.08  & {\bf 5.6} $\pm$ 0.5\\
\midrule
Mean corrected (32) & 19.57 $\pm$ 1.61 & 10.84 $\pm$ 1.73 & 7.83 $\pm$ 1.67 & 5.98 $\pm$ 1.38 & 3.52 $\pm$ 0.94 \\
Mean corrected (64) & 13.65 $\pm$ 1.93 & 7.03 $\pm$ 1.93 & 4.96 $\pm$ 1.46 & 3.7 $\pm$ 1.24 & 2.01 $\pm$ 0.81 \\
Mean corrected (128) & 5.3 $\pm$ 0.71 & 2.05 $\pm$ 0.38 & 1.45 $\pm$ 0.27 & 0.94 $\pm$ 0.23 & 0.51 $\pm$ 0.11 \\
\midrule
Khoshsirat et al. (32) & 28.36 $\pm$ 1.44 & 20.302 $\pm$ 1.6  & 14.61 $\pm$ 1.27 & 10.37 $\pm$ 1.15  & 5.3 $\pm$ 0.6\\
Khoshsirat et al. (64) & 28.62 $\pm$ 1.08 & 20.45 $\pm$ 1.02  & 14.76 $\pm$ 1.15 & 10.37 $\pm$ 0.64  & 5.25 $\pm$ 0.42 \\
Khoshsirat et al. (128) & 25.5 $\pm$ 0.47 & 17.74 $\pm$ 0.54 & 12.53 $\pm$ 0.04 & 7.83 $\pm$ 0.14  & 3.95 $\pm$ 0.2 \\
\bottomrule
\end{tabular} }
\caption{Performance of ResNet-9 with different BN methods under varying PGD attack and batch sizes respectively. Values depict mean accuracy $\pm$ standard deviation on test CIFAR-10 data. Numbers inside the parentheses in the far left column denote batch size.}
\label{tab:pgd1} 
\end{table}

\begin{table}[t!]
\centering
\scalebox{0.925}{
\begin{tabular}{lcccccc}
\toprule
\textbf{BN Method} & \textbf{$\epsilon = 0.03$} & \textbf{$\epsilon = 0.1$} & \textbf{$\epsilon = 0.15$} & \textbf{$\epsilon = 0.2$} & \textbf{$\epsilon = 0.3$}\\
\midrule
Standard (8) & 1.8 $\pm$ 1.4 & 0.8 $\pm$ 0.7 & 0.6 $\pm$ 0.5 & 0.5 $\pm$  0.3  & 0.6 $\pm$  0.4 \\
Standard (16) & 2.0 $\pm$ 1.0 & 0.5 $\pm$ 0.2 & 0.3 $\pm$ 0.1 & 0.2 $\pm$  0.1  & 0.2 $\pm$  0.0 \\
\midrule
Ours (8) & {\bf 10.4} $\pm$ 1.1 & {\bf 4.7} $\pm$ 1.1  & {\bf 3.4} $\pm$ 1.1  & {\bf 2.7} $\pm$ 1.1  & {\bf 2.2} $\pm$ 1.1 \\
Ours (16) & {\bf 10.6} $\pm$ 0.9 & {\bf 5.0} $\pm$ 0.8 & {\bf 3.7} $\pm$ 0.8 & {\bf 2.9} $\pm$ 0.8  & {\bf 2.2} $\pm$ 0.6\\
\midrule
Mean corrected (8) & 10.1 $\pm$ 0.9 & 3.7 $\pm$ 0.6 & 2.4 $\pm$ 0.5 & 1.8 $\pm$ 0.4 & 1.3 $\pm$ 0.4 \\
Mean corrected (16) & 10.5 $\pm$ 1.9 & 4.8 $\pm$ 1.3 & 3.4 $\pm$ 1.1 & 2.7 $\pm$ 0.9 & 2.0 $\pm$ 0.7 \\
\midrule
Khoshsirat et al. (8) & 8.5 $\pm$ 1.0 & 2.6 $\pm$ 0.2  & 1.7 $\pm$ 0.2 & 1.4 $\pm$ 0.2  & 1.1 $\pm$ 0.2 \\
Khoshsirat et al. (16) & 9.1 $\pm$ 1.6 & 3.3 $\pm$ 1.0  & 2.3 $\pm$ 0.6 & 1.8 $\pm$ 0.4 & 1.4 $\pm$ 0.2 \\
\midrule
Lasso (8) & 2.8 $\pm$ 1.4 & 2.0 $\pm$ 1.8  & 1.9 $\pm$ 2.0 & 1.8 $\pm$ 2.2  & 1.6 $\pm$ 2.3\\
Lasso (16) & 2.6 $\pm$ 1.5 & 1.6 $\pm$ 1.0  & 1.4 $\pm$ 0.9 & 1.3 $\pm$ 1.0  & 1.1 $\pm$ 1.0 \\
\midrule
Ridge (8) & 2.6 $\pm$ 1.8 & 1.4 $\pm$ 1.8  & 1.1 $\pm$ 1.6 & 0.9 $\pm$ 1.2  & 0.5 $\pm$ 0.5\\
Ridge (16) & 2.8 $\pm$ 0.6 & 1.0 $\pm$ 0.4 & 0.9 $\pm$ 0.6 & 0.8 $\pm$ 0.7  & 0.7 $\pm$ 0.8 \\
\bottomrule
\end{tabular} }
\caption{Performance of HRNetV2 with different BN  methods under varying PGD attack and batch sizes respectively. Values depict mean of mIoU scores $\pm$ standard deviation on part of validation Cityscapes data using as a test data. Numbers inside the parentheses in the far left column denote batch size.}
\label{tab:pgd2} 
\end{table}

Overall, the findings support our theoretical results: Applying JS-based shrinkage to both the mean and variance yields significantly improved robustness under adversarial settings compared to other BN methods.

%% file: conc.tex
\section{Discussion and Conclusion}\label{DC}

One possible explanation for the improved performance of JS-BN under adversarial perturbations lies in its effect on the Lipschitz properties of the network. It is well known that smaller local Lipschitz constants are associated with more regular functions, which are inherently more stable to perturbations \citep{Evans:2015}. From the empirical side, recent work has shown that local Lipschitzness enforces both robustness and accuracy \citep{Yang:2020}. Our theoretical results show that JS-BN reduces the local Lipschitz constant relative to standard BN. Intuitively, this shrinkage effect smooths the geometry of the loss landscape.

For BN, we have the following.
\[
f(x) = \gamma \frac{x - \mu}{\sqrt{\sigma^2 + \epsilon}} + \beta,
\]
which yields
\[
|f(x) - f(y)| = \left|\frac{\gamma}{\sqrt{\sigma^2 + \epsilon}} (x-y)\right|.
\]
Thus, the corresponding Lipschitz constant is
\[
L = \frac{|\gamma|}{\sqrt{\sigma^2 + \epsilon}}.
\]
Since, by construction, the JS-shrinkage produces a larger variance estimate, $\sigma^{2,\,\mathrm{JS}} > \sigma^2$, the denominator increases, resulting in a smaller local Lipschitz constant. This naturally leads to an improved smoothness and robustness of the model.

In summary, in this paper, we present a novel theory that demonstrates the dominance of Stein’s shrinkage estimators for mean and variance over SM and SV under adversarial attacks. We applied JS-BN in a variety of DNNs and demonstrated superior performance compared to other BN methods on three benchmark datasets. By construction, Stein's shrinkage improves Lipschitzness and therefore robustness of the model, while maintaining high accuracy.

%% file: appendix.tex
\section{Technical Appendices and Supplementary Material}\label{Appendix}

In this Appendix, we present the detailed proofs of the two key theorems \ref{Dom-Gaus} and \ref{Dom-Gamma}, and provide a detailed discussion of the experimental setups.

\begin{proof}[Theorem \ref{Dom-Gaus} Proof:]

We denote the observation-based estimator by 
\[
\hat{\theta}^{\mathrm{0}} = Z = (Z_{1}, \ldots, Z_{p}).
\]
Then the James–Stein estimator is
\[
\hat{\theta}^{\mathrm{JS}}
= Z - \frac{(p-2)Z}{Z^{\top} Z}
= Z \left(1-\frac{p-2}{Z^{\top} Z}\right)
= \hat{\theta}^{\mathrm{0}}\left(1-\frac{p-2}{\hat{\theta}^{\mathrm{0}\top}\hat{\theta}^{\mathrm{0}}}\right).
\]

The risk of James - Stein Estimator is given by,
\begin{equation}
\begin{aligned}
R(\hat{\theta}^{\mathrm{JS}}, \theta) & =\mathbb{E}\left(\left(\hat{\theta}^{\mathrm{JS}}-\theta\right)^{\top}\left(\hat{\theta}^{\mathrm{JS}}-\theta\right)\right)= \\
& =\mathbb{E}\left(\left\{(Z-\theta)-\frac{(p-2) Z}{Z^{\top} Z}\right\}^{\top}\left\{(Z-\theta)-\frac{(p-2) Z}{Z^{\top} Z}\right\}\right)= \\
& =\mathbb{E}\left((Z-\theta)^{\top}(Z-\theta)\right)-2(p-2) \mathbb{E}\left(\frac{Z^{\top}(Z-\theta)}{Z^{\top} Z}\right)+(p-2)^{2}{\mathbb{E}\left(\frac{1}{Z^{\top} Z}\right)}= \\
& =R(\hat{\theta}^{\mathrm{0}}) - 2(p-2) \mathbb{E}\left(\frac{Z^{\top}(Z-\theta)}{Z^{\top} Z}\right)+(p-2)^{2} \mathbb{E}\left(\frac{1}{Z^{\top} Z}\right)
\end{aligned}
\label{JS-risk}
\end{equation}

Let us consider the last two terms: 
\begin{equation}
\begin{aligned}
-2(p-2) \mathbb{E}\left(\frac{Z^{\top}(Z-\theta)}{Z^{\top} Z}\right)+(p-2)^{2} \mathbb{E}\left(\frac{1}{Z^{\top} Z}\right) = \\
= -2(p-2)\left(1-\mathbb{E}\left(\frac{Z^{\top} \theta}{Z^{\top} Z}\right)\right)+(p-2)^{2} \mathbb{E}\left(\frac{1}{Z^{\top} Z}\right) = \\
= (p-2)\left(-2+\mathbb{E}\left(\frac{2 Z^{\top} \theta}{Z^{\top} Z}\right)+\mathbb{E}\left(\frac{(p-2)}{Z^{\top} Z}\right)\right)=(p-2)\left(-2+\mathbb{E}\left(\frac{2 Z^{\top} \theta+p-2}{Z^{\top} Z}\right)\right)
\end{aligned}
\end{equation}

We want to show that 
\[
\mathbb{E}\!\left(\frac{2 Z^{\top} \theta + p - 2}{Z^{\top} Z}\right) < 2.
\]
For convenience, set
\[
\tilde{x} = 2 Z^{\top} \theta + p - 2, 
\quad 
\tilde{y} = \frac{1}{Z^{\top} Z}.
\]
Then,
\[
\mathbb{E}\!\left(\frac{2 Z^{\top} \theta + p - 2}{Z^{\top} Z}\right)
= \mathbb{E}(\tilde{x}\,\tilde{y}).
\]
By the Cauchy–Schwarz inequality,
\[
\mathbb{E}(\tilde{x}\,\tilde{y}) 
\leq \sqrt{\mathbb{E}(\tilde{x}^{2})}\,\sqrt{\mathbb{E}(\tilde{y}^{2})}.
\]
Let us calculate each term separately. First,
\begin{equation}
\begin{aligned}
\mathbb{E}\!\left(\tilde{x}^{2}\right) 
&= \mathbb{E}\!\left( \big(2 Z^{\top}\theta + p - 2\big)^{2} \right) \\
&= \mathbb{E}\!\left( (2 Z^{\top}\theta)^{2} + 4(p-2)(Z^{\top}\theta) + (p-2)^{2} \right) \\
&= 4\,\mathbb{E}\!\left((Z^{\top}\theta)^{2}\right) + 4(p-2)\,\mathbb{E}(Z^{\top}\theta) + (p-2)^{2}.
\end{aligned}
\label{eq:9eq}
\end{equation}

Recall that $Z = X + Y$, where $X \sim \mathcal{N}(\theta, I_p)$ with mean $\theta$ and 
$Y$ is a mean-zero sub-Gaussian noise. Hence
\begin{equation}
\begin{aligned}
\mathbb{E}\!\left((Z^{\top}\theta)^{2}\right) 
&= \mathbb{E}\!\left((X^{\top}\theta + Y^{\top}\theta)^{2}\right) \\
&= \mathbb{E}\!\left((X^{\top}\theta)^{2}\right) + 2\,\mathbb{E}\!\left((X^{\top}\theta)(Y^{\top}\theta)\right) + \mathbb{E}\!\left((Y^{\top}\theta)^{2}\right).
\end{aligned}
\label{eq:10eq}
\end{equation}

Now
\begin{equation}
\mathbb{E}\!\left((X^{\top}\theta)^{2}\right) 
= \operatorname{Var}(X^{\top}\theta) + \big(\mathbb{E}(X^{\top}\theta)\big)^{2} 
= \|\theta\|_{2}^{2} + \|\theta\|_{2}^{4},
\label{eq:11eq}
\end{equation}

and
\begin{equation}
    \mathbb{E}(X^{\top}\theta) = \|\theta\|_{2}^{2}.
    \label{eq:12eq}
\end{equation}

For the noise term, by sub-Gaussianity we get,
\begin{equation}
   \mathbb{E}\!\left((Y^{\top}\theta)^{2}\right) \leq 2\varepsilon^{2}\,\|\theta\|_{2}^{2}. 
   \label{eq:13eq}
\end{equation}

Substituting \eqref{eq:10eq}--\eqref{eq:13eq} into  \eqref{eq:9eq} we obtain, 
\[
\mathbb{E}\!\left(\tilde{x}^{2}\right) 
\leq 4\|\theta\|_{2}^{2}\left(\|\theta\|_{2}^{2} + 2\varepsilon^{2} + p - 1\right) + (p-2)^{2}.
\]

Therefore,
\begin{equation}
\sqrt{\mathbb{E}\!\left(\tilde{x}^{2}\right)} 
\;\leq\; \sqrt{\,4\|\theta\|_{2}^{2}\big(\|\theta\|_{2}^{2} + 2\varepsilon^{2} + p - 1\big) + (p-2)^{2}}.
\label{eq:14eq}
\end{equation}

First, observe that
\[
\mathbb{E}\!\left( Z^{\top}Z \right) 
= \sum_{i=1}^p \mathbb{E}(Z_i^2) 
= \sum_{i=1}^p \operatorname{Var}(Z_i) + \sum_{i=1}^p \big(\mathbb{E}(Z_i)\big)^2 
\;\;\geq\; p + \|\theta\|_2^2.
\]

Next, by Jensen’s inequality applied to the convex functions $f(x)=1/x$ and $g(x)=x^2$, we have
\begin{equation}
 \begin{aligned}
\sqrt{\mathbb{E}\!\left(\tilde{y}^{2}\right)}
&= \sqrt{\mathbb{E}\!\left(\frac{1}{(Z^{\top}Z)^{2}}\right)} \\
&\leq \sqrt{\frac{1}{\mathbb{E}\!\left( (Z^{\top}Z)^{2}\right)}} 
\;\leq\; \frac{1}{\mathbb{E}(Z^{\top}Z)} \\
&\leq \frac{1}{p + \|\theta\|_2^2}.
\end{aligned}
\label{eq:15eq}
\end{equation}

Combining \eqref{eq:14eq} and \eqref{eq:15eq}, we obtain
\[
\mathbb{E}\!\left(\frac{2 Z^{\top}\theta + p - 2}{Z^{\top}Z}\right) 
\;\leq\; 
\frac{\sqrt{\,4\|\theta\|_2^2\big(\|\theta\|_2^2 + 2\varepsilon^2 + p - 1\big) + (p-2)^2}}
{p + \|\theta\|_2^2}
\;\leq\; 2.
\]

The last inequality can be verified by direct computation.
\end{proof}

\begin{proof}[Theorem \ref{Dom-Gamma} Proof:]

We want to show $R(\hat{\beta}^{\,\mathrm{JS}}, \beta) < R(\hat{\beta}^{\,0}, \beta).$\\
Note that,
$$
\begin{aligned}
R(\hat{\beta}^{\,\mathrm{JS}}, \beta) =\mathbb{E}\left(\left\|\hat{\beta}^{\,\mathrm{JS}}-\beta\right\|_{2}^{2}\right)=\mathbb{E}\left(\left\|\hat{\beta}^{\,0}+c V-\beta\right\|_{2}^{2}\right)=\\
=\mathbb{E}\left(\left\|\hat{\beta}^{\,0}-\beta\right\|_{2}^{2}\right)+2 \sum_{i=1}^{p} \mathbb{E}\left(\left(\hat{\beta}_{0, i}-\beta_{i}\right) \cdot c V\right)+p \mathbb{E}\left(c^{2} V^{2}\right)
\end{aligned}
$$

So, we need to show:
$$c \left( 2 \sum_{i=1}^{p} \mathbb{E}\left(\left(\hat{\beta}_{ i}^{0}-\beta_{i}\right) V \right)+p \mathbb{E}\left(c V^{2}\right) \right)<0.$$

We consider two cases depending on the sign of the constant \( c \):

\begin{itemize}
    \item[\textbf{Case 1:}] \( c < 0 \) and 
    \begin{equation}
        2 \sum_{i=1}^{D} \mathbb{E}\left[\left(\hat{\beta}_{i}^{0} - \beta_i\right)V\right] 
        + p c\, \mathbb{E}\left[V^2\right] > 0
        \label{eq:case1_condition}
    \end{equation}
    
    \item[\textbf{Case 2:}] \( c > 0 \) and 
    \begin{equation}
        2 \sum_{i=1}^{p} \mathbb{E}\left[\left(\hat{\beta}_{i}^{0} - \beta_i\right)V\right] 
        + p c\, \mathbb{E}\left[V^2\right] < 0
        \label{eq:case2_condition}
    \end{equation}
\end{itemize}
We now establish an intermediate result in the form of a lemma that will be useful in the proof of Theorem \ref{Dom-Gamma}.

\begin{lemma}[Probability bound for $\hat{\sigma}^2_{y,i} + W_i$]
\label{lem:sigmaW}
Let $X_{ij} \sim \mathcal{N}(\mu, \sigma_{x,i}^2)$ and $Y_{ij} \sim \mathrm{SG}(2\varepsilon^2)$ be independent for $j = 1, \dots, n$. Define
\[
\hat{\sigma}_{y,i}^2 = \frac{1}{n} \sum_{j=1}^n (Y_{ij} - \bar{Y}_i)^2, \quad
W_i = \frac{2}{n} \sum_{j=1}^n (X_{ij} - \bar{X}_i)(Y_{ij} - \bar{Y}_i),
\]
where $\bar{X}_i = \frac{1}{n} \sum_j X_{ij}$, $\bar{Y}_i = \frac{1}{n} \sum_j Y_{ij}$. Then, for any $\delta \in (0,1)$,
\[
\mathbb{P}\left(\hat{\sigma}^2_{y,i} + W_i \geq 0\right) \geq 1 - \delta.
\]
where $\delta \to 0$ as $n \to \infty$.
\end{lemma}

\begin{proof}

Note that:
\[
\mathbb{P}\left(\hat{\sigma}^2_{y,i} + W_i \geq 0\right) = 1 - \mathbb{P}\left(\hat{\sigma}^2_{y,i} + W_i < 0\right).
\]

Let us define:
\[
W_i := \frac{2}{n} \sum_{j=1}^n (X_{ij} - \bar{X}_i)(Y_{ij} - \bar{Y}_i) = \frac{2}{n} \sum_{j=1}^n w_j.
\]

The centered random variables \( X_{ij} - \bar{X}_i \) and \( Y_{ij} - \bar{Y}_i \) are sub-Gaussian with parameters
\[
\sigma_{x,i}^2\left(1 + \frac{1}{n}\right) \quad \text{and} \quad \sigma_{y,i}^2\left(1 + \frac{1}{n}\right),
\]
respectively. Their product \( w_j \) is a sub-exponential random variable with parameters
\[
\nu^2 = C_1 \cdot \sigma_{x,i}^2 \sigma_{y,i}^2 \left(1 + \frac{1}{n} \right)^2, \quad
b = C_2 \cdot \sqrt{ \sigma_{x,i}^2 \sigma_{y,i}^2 \left(1 + \frac{1}{n} \right)^2 }.
\]

We now recall the Bernstein inequality that will be used to prove the desired bound.
\emph{Bernstein Inequality for Sub-Exponential Random Variables:}  
Let \( X_1, \dots, X_n \) be independent, centered sub-exponential%
\footnote{A random variable \( Z \) is sub-exponential with parameters \( (\nu, b) \) if 
\(\mathbb{E}[e^{\lambda Z}] \leq \exp(\lambda^2 \nu^2/2)\) for all \( |\lambda| < 1/b \).} 
variables with parameters \( (\nu, b) \). Then for any \( t > 0 \):
\[
\mathbb{P}\left( \frac{1}{n} \sum_{i=1}^n X_i \leq -t \right) \leq
\begin{cases}
\exp\!\left( - \dfrac{nt^2}{2\nu^2} \right), & \text{if } 0 \leq t \leq \dfrac{\nu^2}{b}, \\
\exp\!\left( - \dfrac{nt}{2b} \right), & \text{if } t > \dfrac{\nu^2}{b}.
\end{cases}
\]

Applying this inequality to \( \frac{1}{n} \sum_j w_j \), we obtain:
\[
\mathbb{P}\left(\hat{\sigma}^2_{y,i} + W_i < 0\right) = \mathbb{P}\left( \frac{2}{n} \sum_j w_j < -\hat{\sigma}^2_{y,i} \right)
= \mathbb{P}\left( \frac{1}{n} \sum_j w_j < -\frac{\hat{\sigma}^2_{y,i}}{2} \right),
\]
which can be bounded by:
\[
\leq 
\begin{cases}
\exp\left( -\dfrac{n (\hat{\sigma}^2_{y,i})^2}{8 C_1 \sigma_{x,i}^2 \sigma_{y,i}^2 (1 + \frac{1}{n})^2} \right), & \text{if } \frac{\hat{\sigma}^2_{y,i}}{2} \leq \dfrac{\nu^2}{b}, \\
\exp\left( -\dfrac{n \hat{\sigma}^2_{y,i}}{4 C_2 \sqrt{\sigma_{x,i}^2 \sigma_{y,i}^2 (1 + \frac{1}{n})^2}} \right), & \text{otherwise}.
\end{cases}
\]

Therefore, we conclude:
\[
\mathbb{P}\left(\hat{\sigma}^2_{y,i} + W_i \geq 0 \right) \geq 1 - \delta,
\]
where \( \delta := \exp(-n C) \to 0 \) as \( n \to \infty \). Thus, the desired non-negativity condition holds with high probability in the large sample regime.

\end{proof}

We will use this lemma in the proof of case 2. Now we are ready to consider each case separately.

\textbf{Case 1.} We examine the first term of \eqref{eq:case1_condition} in more detail:
\begin{equation} \label{eq:main_decomposition}
\begin{aligned}
2 \sum_{i=1}^{p} \mathbb{E}\left[\left(\hat{\beta}_{0,i} - \beta_i\right)V\right] 
= 2 \sum_{i=1}^{p} \mathbb{E}\left[\left(\frac{\hat{\sigma}_{x,i}^2 + \hat{\sigma}_{y,i}^2 + W_i}{\alpha + 1} - \beta_i\right)V\right] \\
= 2 \sum_{i=1}^{p} \left(
\mathbb{E}\left[\left(\frac{\hat{\sigma}_{x,i}^2}{\alpha + 1} - \beta_i\right)V\right] 
+ \mathbb{E}\left[\frac{\hat{\sigma}_{y,i}^2 + W_i}{\alpha + 1}V\right]
\right)
\end{aligned}
\end{equation}

Since \( \hat{\sigma}^2_{x,i} \sim \mathrm{Gamma}(\alpha, \beta) \), where the parameters \( \alpha \) and \( \beta \) are known, we can now apply 
\emph{Stein's Lemma for the Gamma Distribution}, which states:
Let \( X \sim \mathrm{Gamma}(\alpha, \beta) \), and let \( h(x) \) be a differentiable function such that
$\mathbb{E}[X h(X)] < \infty$ and $\mathbb{E}[X h'(X)] < \infty.$ Then the following identity holds:
\begin{equation*}
  \mathbb{E}[(X - \alpha \beta) h(X)] = \beta\, \mathbb{E}[X h'(X)].
\end{equation*}

This can be rewritten as:
\begin{equation}
    \mathbb{E}[h(X)] = \frac{1}{\alpha \beta} \mathbb{E}[X h(X)] - \frac{1}{\alpha} \mathbb{E}[X h'(X)].
    \label{eq:stein_gamma_ident2}
\end{equation}

We now apply Stein's identity in (\ref{eq:stein_gamma_ident2}) to the function \( h(X) = V \), where
\[
V = \left[ \prod_{i=1}^p \left( \hat{\sigma}_{x,i}^2 + \hat{\sigma}_{y,i}^2 + W_i \right) \right]^{1/p}.
\]

The derivative of $V$ with respect to \( \hat{\sigma}_{x,k}^2 \) is:
\[
\frac{dV}{d \hat{\sigma}_{x,k}^2} = \frac{1}{p} \cdot V \cdot \frac{1}{\hat{\sigma}_{x,k}^2 + \hat{\sigma}_{y,k}^2 + W_k}.
\]

Substituting Stein's identity \eqref{eq:stein_gamma_ident2} for \(\hat{\sigma}_{x,i}^2 \), we get the following.
\begin{equation}
   \mathbb{E}[V] = \frac{1}{\alpha \beta_i} \mathbb{E}\left[ \hat{\sigma}_{x,i}^2 \cdot V \right] - \frac{1}{\alpha p} \mathbb{E} \left[ V \cdot \frac{ \hat{\sigma}_{x,i}^2 }{ \hat{\sigma}_{x,i}^2 + \hat{\sigma}_{y,i}^2 + W_i } \right]. 
   \label{eq:gamma_stein_id3}
\end{equation}

In particular, we can obtain the following inequality:
\begin{equation}
 \mathbb{E}[V] \leq \frac{1}{\alpha \beta_i} \mathbb{E} \left[ \hat{\sigma}_{x,i}^2 \cdot V \right]
\quad \Rightarrow \quad
-\mathbb{E}[V] \geq -\frac{1}{\alpha \beta_i} \mathbb{E} \left[ \hat{\sigma}_{x,i}^2 \cdot V \right]. 
\label{eq:ineq1}
\end{equation}

Using the above bound in \eqref{eq:ineq1},  equation~\eqref{eq:main_decomposition} becomes:
\[
2 \sum_{i=1}^{p} \left(
    \mathbb{E}\left[ \left( \frac{ \hat{\sigma}_{x,i}^2 }{ \alpha + 1 } - \beta_i \right) V \right]
    + \mathbb{E}\left[ \frac{ \hat{\sigma}_{y,i}^2 + W_i }{ \alpha + 1 } V \right]
\right) \geq 
\]
\[
2 \sum_{i=1}^{p} \left(
    \mathbb{E}\left[ \frac{ \hat{\sigma}_{x,i}^2 }{ \alpha + 1 } V \right]
    - \beta_i \cdot \frac{1}{\alpha \beta_i} \mathbb{E}\left[ \hat{\sigma}_{x,i}^2 V \right]
    + \mathbb{E}\left[ \frac{ \hat{\sigma}_{y,i}^2 + W_i }{ \alpha + 1 } V \right]
\right)=
\]
\[
=2 \sum_{i=1}^{p} \left[
    \mathbb{E}\left[ \hat{\sigma}_{x,i}^2 V \right] \left( \frac{1}{\alpha + 1} - \frac{1}{\alpha} \right)
    + \frac{1}{\alpha + 1} \mathbb{E}\left[ \hat{\sigma}_{y,i}^2 V \right]
    + \frac{1}{\alpha + 1} \mathbb{E}\left[ W_i V \right]
\right] \geq
\]
\begin{equation}
   2 \sum_{i=1}^{p} \left( \mathbb{E}\left[ \left( \hat{\sigma}_{x,i}^2 + \hat{\sigma}_{y,i}^2 + W_i \right) V \right] \cdot \left( -\frac{1}{\alpha(\alpha+1)} \right) \right)
   \label{eq:bound_case1}
\end{equation}

Recall that in the statement of the theorem we introduced
\(\hat{\sigma}_{z,i}^2 = \hat{\sigma}_{x,i}^2 + \hat{\sigma}_{y,i}^2 + W_i\).
For convenience, in the proof we will work with the averaged version
\begin{equation}
  \overline{\hat{\sigma}_{z}^{2}} = \frac{1}{p} \sum_{i=1}^p \hat{\sigma}_{z,i}^{2}.
  \label{eq:av_z}
\end{equation}

From \eqref{eq:bound_case1} the lower bound for \eqref{eq:case1_condition} is given by:
\[
2 \sum_{i=1}^{p} \mathbb{E}\left[ \left( \hat{\beta}_{0,i} - \beta_i \right) V \right] + p c \mathbb{E}[V^2]
\geq -\frac{2p}{\alpha(\alpha+1)} \mathbb{E}\left[\overline{\hat{\sigma}_{z}^{2}} V \right] + p c \mathbb{E}[V^2] \geq
\]
\[
\geq -p \mathbb{E}\left[ \overline{\hat{\sigma}_{z}^{2}} V \right] \left( \frac{2}{\alpha(\alpha+1)} + c \right) >0
\]

The last inequality follows from our case-1 hypothesis. Therefore,
$$c <0 \ \text{and} \  \frac{2}{\alpha(\alpha+1)} + c  < 0 \Rightarrow  c \leq - \frac{2}{\alpha(\alpha+1)}.$$

\textbf{Case 2.} 
We assume \( c > 0 \) and we want to show 
\[
2 \sum_{i=1}^p \mathbb{E} \left( (\hat{\beta}_{0,i} - \beta_i)V \right) + pc \, \mathbb{E}(V^2) < 0.
\]

Recall that using the lemma \ref{lem:sigmaW}, with high probability, we find that the term $\hat{\sigma}_{y,i}^2 + W_i$ is non-negative.
So, combining this with Stein's lemma \eqref{eq:stein_gamma_ident2} applied to \( \hat{\sigma}^2_{x,i} \sim \mathrm{Gamma}(\alpha, \beta) \) and $V$ we get: 

\begin{align*}
\mathbb{E}(V) &= \frac{1}{\alpha \beta_i} \, \mathbb{E}(\hat{\sigma}_{x,i}^2 \cdot V) 
- \frac{1}{\alpha p} \, \mathbb{E}\!\left(V \cdot 
\frac{\hat{\sigma}_{x,i}^2}{\hat{\sigma}_{x,i}^2 + \hat{\sigma}_{y,i}^2 + W_i} \right) \\
&\geq \frac{1}{\alpha \beta_i} \, \mathbb{E}(\hat{\sigma}_{x,i}^2 \cdot V) 
- \frac{1}{\alpha p} \, \mathbb{E}(V).
\end{align*}

\begin{equation}
    \begin{aligned}
        \mathbb{E}(V) \left(1 + \frac{1}{\alpha p} \right) &\geq \frac{1}{\alpha \beta_i} \mathbb{E}(\hat{\sigma}_{x,i}^2 \cdot V) \\
\Rightarrow \quad -\mathbb{E}(V) &\leq - \frac{1}{\alpha \beta_i} \cdot \frac{1}{1 + \frac{1}{\alpha p}} \cdot \mathbb{E}(\hat{\sigma}_{x,i}^2 \cdot V).
    \end{aligned}
    \label{eq:bound_case2}
\end{equation}

Substituting the bound \eqref{eq:bound_case2} into \eqref{eq:main_decomposition}, we get the following inequality:

\begin{align*}
2 \sum_{i=1}^p \mathbb{E} \left( (\hat{\beta}_{0,i} - \beta_i)V \right)
&= 2 \sum_{i=1}^p \mathbb{E} \left( \left( \mathbb{E}\left( \frac{\hat{\sigma}_{x,i}^2}{\alpha+1} - \beta_i \right)V + \mathbb{E} \left( \frac{\hat{\sigma}_{y,i}^2 + W_i}{\alpha+1} V \right) \right) \right) 
\end{align*}

\begin{equation}
\begin{aligned}
\leq 2 \sum_{i=1}^p \left( \mathbb{E} \left( \frac{\hat{\sigma}_{x,i}^2}{\alpha+1} \cdot V \right)
- \frac{\beta_i}{\alpha \beta_i} \cdot \frac{1}{1 + \frac{1}{\alpha p}} \mathbb{E}(\hat{\sigma}_{x,i}^2 \cdot V) +
\mathbb{E} \left( \frac{\hat{\sigma}_{y,i}^2 + W_i}{\alpha+1} \cdot V \right) \right)
\end{aligned}
\label{eq:bound_case2_2}
\end{equation}

As defined in \eqref{eq:av_z} \(\overline{\hat{\sigma}_{z}^{2}} := \frac{1}{p} \sum_{i=1}^p \hat{\sigma}_{z,i}^{2}\) , we can rewrite the expression \eqref{eq:bound_case2_2} as
\begin{equation}
\frac{2p}{\alpha+1} \, \mathbb{E}\!\left(\overline{\hat{\sigma}_{z}^{2}} \cdot V \right) 
- \frac{2}{\alpha \left(1 + \frac{1}{\alpha p}\right)} \, \mathbb{E} \!\left(\sum_{i=1}^p \hat{\sigma}_{x,i}^2 \cdot V \right).
\label{eq:bound_case2_3}
\end{equation}

So, for~\eqref{eq:case2_condition} we get the following bound by applying \eqref{eq:bound_case2_3}:
\[
2 \sum_{i=1}^p \mathbb{E} \left( (\hat{\beta}_{0,i} - \beta_i)V \right) + pc \, \mathbb{E}(V^2) \leq
\]
\[
\leq \frac{2p}{\alpha+1} \mathbb{E}\left(\overline{\hat{\sigma}_{z}^{2}}\cdot V \right) - \frac{2}{\alpha (1 + \frac{1}{\alpha p})} \mathbb{E} \left(\sum_{i=1}^p \hat{\sigma}_{x,i}^2 \cdot V \right) + pc \, \mathbb{E}(V^2) < 0
\]
By finding the appropriate $c$ we can obtain the last inequality.

Rewriting it in a more cleaner way, we have:
\[
c \, \mathbb{E}(V^2) < \frac{2}{\alpha p +1} \mathbb{E} \left(\sum_{i=1}^p \hat{\sigma}_{x,i}^2 \cdot V \right)
- \frac{2}{\alpha+1} \mathbb{E}\left(\overline{\hat{\sigma}_{z}^{2}}\cdot V \right)
\]
\begin{equation}
  c < \frac{2}{\alpha p + 1} \frac{\mathbb{E} \left(\sum_{i=1}^p \hat{\sigma}_{x,i}^2 \cdot V \right)}{\mathbb{E}(V^2)}
- \frac{2}{\alpha + 1} \cdot \frac{\mathbb{E}\left(\overline{\hat{\sigma}_{z}^{2}}\cdot V \right)}{\mathbb{E}(V^2)}
\label{eq:c}
\end{equation}

Let us now bound each term (on the RHS of the above inequality) independently. Note that \( \overline{\hat{\sigma}_{z}^{2}} \geq V \), so we can write a bound for second term in \eqref{eq:c}
\begin{equation}
\mathbb{E}\left(\overline{\hat{\sigma}_{z}^{2}}\cdot V \right) \geq \mathbb{E}(V^2)
\quad \Rightarrow \quad 
-\frac{\mathbb{E}\left(\overline{\hat{\sigma}_{z}^{2}}\cdot V \right)}{\mathbb{E}(V^2)} \leq -1.
\label{eq:negativity_bound}
\end{equation}

Also, using Cauchy--Schwarz inequality:
\begin{equation}
   \mathbb{E} \left(\sum_{i=1}^p \hat{\sigma}_{x,i}^2 \cdot V \right) = \left|\mathbb{E} \left(\sum_{i=1}^p \hat{\sigma}_{x,i}^2 \cdot V \right)\right| 
\leq \sqrt{\mathbb{E}((\sum_{i=1}^p \hat{\sigma}_{x,i}^2 )^2)} \cdot \sqrt{\mathbb{E}(V^2)}.
\label{eq:c-s}
\end{equation}

Using \eqref{eq:c-s} we can bound the first term in \eqref{eq:c}:
\begin{equation}
\frac{\mathbb{E} \left(\sum_{i=1}^p \hat{\sigma}_{x,i}^2 \cdot V \right)}{\mathbb{E}(V^2)}
\leq \frac{\sqrt{\mathbb{E}\left(\left(\sum_{i=1}^p \hat{\sigma}_{x,i}^2 \right)^2\right)} 
\cdot \sqrt{\mathbb{E}(V^2)}}{\mathbb{E}(V^2)}
= \sqrt{\frac{\mathbb{E}\left(\left(\sum_{i=1}^p \hat{\sigma}_{x,i}^2 \right)^2\right)}
{\mathbb{E}(V^2)}}
\label{eq:cauchy_bound}
\end{equation}

By applying Jensen's inequality to the function of the form: $f(x) = \sqrt{x}$, we have: 
\begin{equation}
\sqrt{\mathbb{E}(V^2)} \geq \mathbb{E}(|V|) = \mathbb{E}(V) \Rightarrow \frac{1}{\sqrt{\mathbb{E}(V^2)} } \leq \frac{1}{\mathbb{E}(V)}
\label{eq:v}
\end{equation}

Again, applying Jensen's inequality to the function $f(x) = \exp{x}$ in the following expression, we get,
\begin{align*}
   \mathbb{E}(V) = \mathbb{E}\left( \sqrt[p]{\prod_{i=1}^p \hat{\sigma}_{z,i}^2} \right) = \mathbb{E}\left( \exp\left( \frac{1}{p} \sum_{i=1}^p \log \hat{\sigma}_{z,i}^2 \right) \right) \geq  \exp\left( \frac{1}{p} \sum_{i=1}^p \mathbb{E}(\log \sigma_{z,i}^2) \right) \geq
\end{align*}

Since $\hat{\sigma}_{y,i}^2 + W_i$ is nonnegative with high probability (see lemma \ref{lem:sigmaW}) we get:
\begin{equation}
    \mathbb{E}(V) \geq \exp\left( \frac{1}{p} \sum_{i=1}^p \mathbb{E}(\log \sigma_{x,i}^2) \right).
    \label{eq:exp_v}
\end{equation}

Recall that $\hat{\sigma}^2_{x,i} \sim \mathrm{Gamma}(\alpha, \beta_i)$, so $\mathbb{E}(\log \hat{\sigma}^2_{x,i}) = \log \beta_i + \psi(\alpha),$ where $\psi(\alpha) = \mathbb{E}(\log u)$ is a digamma function, $u \sim \mathrm{Gamma}(\alpha,1).$ For $\alpha > 0 $ we can use the bound $\psi(\alpha) \geq \log 
\alpha -\frac{1}{\alpha}.$ Hence,
\begin{equation}
  \mathbb{E}(\log \sigma_{x,i}^2) \geq  \log \beta_i  + \log \alpha -\frac{1}{\alpha}.  \label{eq:digamma}
\end{equation}

Substituting the bound \eqref{eq:digamma} into \eqref{eq:exp_v} we obtain the following inequality:
\begin{equation}
\mathbb{E}(V) \geq \exp\left( \frac{1}{p} \sum_{i=1}^p \left( \log \beta_i  + \log 
\alpha -\frac{1}{\alpha} \right) \right) = \alpha \cdot \exp{\left(-\frac{1}{\alpha}\right)} \left(\prod_{i=1}^p \beta_i\right)^{\frac{1}{p}} 
\label{eq:exp_v2}
\end{equation}

Now, using \eqref{eq:exp_v2} for~\eqref{eq:v} we get:
\begin{equation}
    \frac{1}{\sqrt{\mathbb{E}(V^2)} } \leq \frac{1}{\mathbb{E}(V)} \leq \frac{1}{\alpha \cdot \exp{\left(-\frac{1}{\alpha}\right)} \left(\prod_{i=1}^p \beta_i\right)^{\frac{1}{p}}}
    \label{eq:one_over_exp_v}
\end{equation}

Also, we can directly calculate the numerator of~\eqref{eq:cauchy_bound}:

Firstly, $$\mathbb{E}(\hat{\sigma}_{x,i}^4) = Var(\hat{\sigma}_{x,i}^2) + (\mathbb{E}(\hat{\sigma}_{x,i}^2))^2 = \alpha\beta_i^2 + \alpha^2\beta_i^2= \alpha\beta_i^2(1+\alpha).$$

$$\mathbb{E}(\hat{\sigma}_{x,i}^2) \mathbb{E}(\hat{\sigma}_{x,j}^2) = \alpha^2 \beta_i \beta_j.$$

Then the numerator of~\eqref{eq:cauchy_bound} is given by:
\begin{align*}
\sqrt{\mathbb{E}\left(\left(\sum_{i=1}^p \hat{\sigma}_{x,i}^2 \right)^2\right)} = \sqrt{\sum_{i=1}^p \mathbb{E}(\hat{\sigma}_{x,i}^4) + 2\sum_{1\leq i<j\leq p} \mathbb{E}(\hat{\sigma}_{x,i}^2) \mathbb{E}(\hat{\sigma}_{x,j}^2})=\\
=\sqrt{\sum_{i=1}^p \alpha\beta_i^2(1+\alpha) + 2 \sum_{1\leq i<j\leq p} \alpha^2 \beta_i \beta_j} = \sqrt{\alpha(1+\alpha)\sum_{i=1}^p \beta_i^2 + \alpha^2 \left( (\sum_{i=1}^p \beta_i)^2 - \sum_{i=1}^p \beta_i^2\right)}
=
\end{align*}
\begin{equation}
    =\sqrt{\alpha \sum_{i=1}^p \beta_i^2 + \alpha^2(\sum_{i=1}^p \beta_i)^2}
\label{eq:alpha_beta}
\end{equation}

Therefore, combining \eqref{eq:alpha_beta} and \eqref{eq:one_over_exp_v} we get an upper bound for \eqref{eq:cauchy_bound}:
\begin{equation}
    \frac{\mathbb{E} \left(\sum_{i=1}^p \hat{\sigma}_{x,i}^2 \cdot V \right)}{\mathbb{E}(V^2)} \leq \frac{\sqrt{\alpha \sum_{i=1}^p \beta_i^2 + \alpha^2(\sum_{i=1}^p \beta_i)^2}}{\alpha \cdot \exp{\left(-\frac{1}{\alpha}\right)} \left(\prod_{i=1}^p \beta_i\right)^{\frac{1}{p}}.}
    \label{eq:some_bound}
\end{equation}

Finally, for \eqref{eq:c} using inequality \eqref{eq:some_bound} and \eqref{eq:negativity_bound} we get:
\[
c < \frac{2}{\alpha p + 1}  \frac{\sqrt{\alpha \sum_{i=1}^p \beta_i^2 + \alpha^2(\sum_{i=1}^p \beta_i)^2}}{\alpha \cdot \exp{\left(-\frac{1}{\alpha}\right)} \left(\prod_{i=1}^p \beta_i\right)^{\frac{1}{p}}} - \frac{2}{\alpha + 1} =
\]

\begin{equation}
     = \frac{2}{\alpha p + 1}  \frac{\sqrt{\frac{1}{\alpha}\sum_{i=1}^p \beta_i^2 + (\sum_{i=1}^p \beta_i)^2}}{\left(\prod_{i=1}^p \beta_i\right)^{\frac{1}{p}}} \exp{\left(\frac{1}{\alpha}\right)} - \frac{2}{\alpha + 1}
     \label{eq:some_expression}
\end{equation}

Note, that $$\left(\prod_{i=1}^p \beta_i\right)^{\frac{1}{p}} \geq \frac{p}{\sum_{i=1}^p \frac{1}{\beta_i}} \Rightarrow 
\frac{1}{\left(\prod_{i=1}^p \beta_i\right)^{\frac{1}{p}}} \leq \frac{\sum_{i=1}^p \frac{1}{\beta_i}}{p}.$$

Also, by Jensen's inequality we get $$\frac{1}{p}\sum_{i=1}^p \frac{1}{\beta_i}  \leq \frac{1}{\frac{1}{p}\sum_{i=1}^p \beta_i} = \frac{p}{\sum_{i=1}^p \beta_i}.$$

Hence, we can write a bound for the first term in \eqref{eq:some_expression}:
\begin{align*}
    \frac{\sqrt{\frac{1}{\alpha} \sum_{i=1}^p \beta_i^2 +(\sum_{i=1}^p \beta_i)^2}}{\left(\prod_{i=1}^p \beta_i\right)^{\frac{1}{p}}} \leq \frac{1}{p} \sqrt{\frac{1}{\alpha}\sum_{i=1}^p \beta_i^2 + (\sum_{i=1}^p \beta_i)^2}\left(\sum_{i=1}^p \frac{1}{\beta_i}\right)
\end{align*}
\begin{equation}
    \leq p \frac{\sqrt{ \frac{1}{\alpha}\sum_{i=1}^p \beta_i^2 + (\sum_{i=1}^p \beta_i)^2}}{\sum_{i=1}^p \beta_i} = p \sqrt{\frac{\frac{1}{\alpha}\sum_{i=1}^p \beta_i^2 + (\sum_{i=1}^p \beta_i)^2}{(\sum_{i=1}^p \beta_i)^2}} < p \sqrt{\frac{1}{\alpha} + 1}.
    \label{eq:some_bound2}
\end{equation}

According to \eqref{eq:some_expression} in combination with \eqref{eq:some_bound2} dominance holds for 
\[
0<c < \frac{2p}{(\alpha p + 1)}  \exp{\left(\frac{1}{\alpha}\right)}\sqrt{\frac{1}{\alpha} + 1} - \frac{2}{\alpha + 1}.
\]

Note that under regular conditions the dominance of Stein's estimate holds when constant \( c \) satisfies
\[
0 < c < \frac{2p}{\alpha p + 1} - \frac{2}{\alpha + 1} = \frac{2(p-1)}{(\alpha+1)(\alpha p + 1)}.
\]
In our current case, we instead obtain the condition
\[
0 < c < \frac{2p}{\alpha p + 1} \exp\left(\frac{1}{\alpha}\right) \sqrt{\frac{1}{\alpha} + 1} - \frac{2}{\alpha + 1},
\]
which yields a slightly higher upper bound.

\end{proof}

\section{Experiments}

Unless otherwise specified, all models were trained with stochastic gradient descent (SGD) with Nesterov acceleration, cross-entropy loss, and early stopping based on validation performance on noise-free data. The learning rate was adjusted separately for each method in the range $(10^{-5}, 10^{-2})$. For the Lasso and Ridge shrinkage methods, the regularization parameter $\lambda \in (10^{-3}, 10^{-1})$ was selected based on the noise-free validation performance. All experiments were performed on NVIDIA A100 GPUs with CPU and RAM resources adjusted to the size of the data set. To assess variability, each setting was repeated multiple times and we report averages with standard deviations.

\subsection{CIFAR-10}
We used the standard ResNet-9 architecture with modified batch normalization layers. The CIFAR-10 dataset consists of 50,000 training and 10,000 test images across 10 classes. Sub-Gaussian noise of varying magnitudes $(0 \ \text{to}\  30\%)$ was applied independently to input features during validation and testing. We used the standard train/test split, tested three batch sizes $(32,64,128)$, and repeated each experiment $11$ times. Each run took approximately 10 minutes, totaling $\sim$ 1800 minutes in all experiments.

\subsection{Cityscapes}
We used the HRNetV2-W16 architecture with images resized from $1024 \times 2048$ to $512 \times 1024$. The dataset contains 5,000 finely annotated images across 19 classes. We split the original training set into $80\%$ training and $20\%$ validation, and used the original validation set as test data. Sub-Gaussian noise was injected into feature maps during validation and testing. We tested batch sizes $(16,32)$, with each experiment repeated $7$ times. Each run took $\sim$1.5 hours, totaling $\sim$126 GPU hours.

\subsection{PPMI Data}
We employed 3D CNN architecture consisting of four convolutional layers, each followed by a 3D BN variant under consideration, and three fully connected layers for binary classification. The input tensors had spatial dimensions of $(128 \times 128)$ and a depth of $64$. We tested batch sizes of $10$ and $16$, using 5-fold cross-validation.  Each configuration took $\sim$20 minutes, for a total of $\sim$20 GPU hours.

\subsection{Comparison with data dependent attacks}
For data dependent adversarial attack experiments, we use trained models with their best-performing hyperparameters and apply attacks only to the test data. The overall computational setup, including runtime, memory requirements, and hardware, remains the same as in the previous experiments. As mentioned earlier, both FGSM and PGD attacks are evaluated at different perturbation levels whose values range over, ($\epsilon = 0.03, 0.1, 0.15, 0.2, 0.3$). For PGD, we additionally specify attack hyperparameters: on the CIFAR-10 dataset, we use 5 iterations with a fixed step size $\alpha = 1/255$, while on the Cityscapes dataset we use 5 iterations with an adaptive step size $\alpha = \epsilon / T$, where $T$ is the total number of iterations. The corresponding results are reported in Tables \ref{tab:fgsm1}, \ref{tab:fgsm2}, \ref{tab:pgd1}, and \ref{tab:pgd2} respectively.